\documentclass[runningheads]{llncs}

\usepackage{subfiles} 
\usepackage{subfigure}
\usepackage{graphicx}
\usepackage{csquotes}
\usepackage{amsmath}
\usepackage{algorithm}
\usepackage[noend]{algpseudocode}
\usepackage{amssymb}
\usepackage[show]{chato-notes}
\usepackage{comment}
\usepackage{booktabs}
\usepackage{multirow}
\usepackage{multicol}
\usepackage{wrapfig}
\usepackage{cleveref}
\usepackage{xcolor}
\usepackage{xurl}

\newcommand{\spara}[1]{\smallskip\noindent\textbf{#1}}
\newcommand{\mpara}[1]{\medskip\noindent\textbf{#1}}

\newcommand{\dcs}{\texttt{DCS}\xspace}
\newcommand{\tri}{\texttt{TRI}\xspace}
\newcommand{\cli}{\texttt{CLI}\xspace}
\newcommand{\rcli}{\texttt{RCLI}\xspace}
\newcommand{\edg}{\texttt{EDG}\xspace}
\newcommand{\dataset}{\texttt{DAT}\xspace}
\newcommand{\datasetbw}{\texttt{DAT+BW}\xspace}
\newcommand{\rclibw}{\texttt{RCLI+BW}\xspace}

\begin{document}

\title{Counterfactual Explanations for Graph Classification Through the Lenses of Density}

\titlerunning{Counterfactual Explanations for Graph Classification}

\author{Carlo Abrate\inst{1}\orcidID{0009-0003-8604-9699} \and
Giulia Preti\inst{1}\orcidID{0000-0002-2126-326X} \and
Francesco Bonchi\inst{1,2}\orcidID{0000-0001-9464-8315}}

\authorrunning{C. Abrate et al.}

\institute{
    CENTAI Institute, Turin, Italy \and
    EURECAT, Barcelona, Spain\\
    \email{\{carlo.abrate,giulia.preti,bonchi\}@centai.eu}
}

\maketitle \sloppy              

\begin{abstract}
Counterfactual examples have emerged as an effective approach to produce simple and understandable post-hoc explanations. In the context of graph classification,
previous work has focused on generating counterfactual explanations by manipulating the most elementary units of a graph, i.e., removing an existing edge, or adding a non-existing one. In this paper, we claim that such language of explanation might be too fine-grained, and turn our attention to some of the main characterizing features of real-world complex networks, such as the tendency to close triangles, the existence of recurring motifs, and the organization into dense modules. We thus define a general
\emph{density-based counterfactual search} framework to generate instance-level counterfactual explanations for graph classifiers, which can be instantiated with different notions of dense substructures. In particular, we show two specific instantiations of this general framework: a method that searches for counterfactual graphs by opening or closing triangles, and a method driven by maximal cliques.
We also discuss how the general method can be instantiated to exploit any other notion of dense substructures, including, for instance, a given taxonomy of nodes.
We evaluate the effectiveness of our approaches in 7 brain network datasets and compare the counterfactual statements generated according to several widely-used metrics. Results confirm that adopting a semantic-relevant unit of change like density is essential to define versatile and interpretable counterfactual explanation methods.

\end{abstract}
\section{Introduction}\label{sec:intro}
Graphs provide a flexible, expressive, and powerful data representation paradigm to model complex systems made of entities and relationships between them, such as users in social networks, regions in the brain, and proteins in an organism.
A widely studied task on graph-structured data is \emph{graph classification}, which involves assigning labels or categories to graphs based on their structural properties or node features.
Graph classification has benefited greatly from the many recent technical advances, especially thanks to \emph{graph neural networks} (GNN). However, as AI techniques become more complex, it becomes challenging to understand their output~\cite{DBLP:journals/corr/abs-2012-15445}.
This opacity can lead to uninformed decisions, complicate the audit process, and ultimately limit the trust in AI techniques, and thus their adoption.
In these regards, post-hoc explanation methods have emerged as an approach to make black-box models more interpretable \cite{biran2017explanation,guidotti2018survey}.
Explanations of black-box models can help to build trust in AI systems by enabling users to understand the decision-making process and assess the reliability of the system.
While this is particularly important in applications that impact people's lives, such as healthcare, finance, and justice~\cite{bhatore2020machine,kononenko2001machine}, explaining AI models is also
of uttermost importance in biological domains in which, more than the mere classification accuracy, it is important for the scientist to understand which modules play a role in a specific pathology or biological condition. For instance, in brain networks analysis, the neuroscientist needs to understand which are the
regions of the brain whose interactions discriminate between
disordered and healthy individuals~\cite{ha2015characteristics,weston2019four}.

\textit{Counterfactual explanations}~\cite{Wachter2017,moraffah2020causal} are a method
for providing post-hoc explanations of individual instance classification.
These explanations consist in a counterfactual example, which is a modified version of the instance that leads to a different classification.
They take the form of a counterfactual statement, such as \emph{``If X had been different, Y would not have occurred"}, which is typically concise and easy to understand~\cite{karimi2020survey}.
Defining and generating optimal counterfactuals is a challenging task, especially when working with graph data. This is due to the large space of interdependent features
and the complex interconnections between the nodes, which can make determining which features to modify difficult and time-consuming. Additionally, some states within the feature space may be too complex for humans to fully comprehend or may be difficult to explain using the same semantic associated with the type of data under consideration.

\spara{Graph classification for brain networks.}
In this work, without loss of generality, we adopt as the main application example, the binary classification of brain networks~\cite{wang2017brainclassification,du2018brainclassificationsurvey,meng2018brainclassification,yan2019groupinn,misman2019brainclassification,cslanciano}.
Brain networks can be modeled as undirected graphs, with nodes denoting \emph{regions of interest} (ROIs), and edges indicating correlations of activation.
In brain networks classification we are given two groups of individuals, e.g.,
a condition group  and
a control group, where each individual is represented by a graph $G_i=(V,E_i)$,
defined over the same set $V$ of nodes (corresponding to the ROIs).
The set of edges $E_i$ represents the connections, either structural or functional, between the ROIs of the observed $G_i$.
The goal is to learn a binary classifier $f: \mathcal{G} \rightarrow \{0, 1\}$
which, given an unseen brain network $G_n=(V,E_n)$,
predicts to which of the two groups it belongs.

Besides brain networks, this specific type of graph classification task, i.e., graph classification \emph{with node identity awareness} \cite{Barabsi2004,gutierrez2019embedding,Koutrouli_Karatzas_2020,you2021identity}, occurs whenever the identity of the node is an important information which identifies the same entity across all the input graphs. This is, for instance, the case
in \emph{``omics''} domains, such as in gene co-expression networks
\cite{bth379,giad010}, protein-protein interaction networks \cite{wsbm121,gulfidan2020pan}, or gene regulatory networks \cite{kim2018diffgrn,singh2018differential}.

\spara{Density-based graph counterfactuals.}
Intuitively, given a specific graph $G$ and a binary classifier $f$, a \emph{counterfactual graph}~\cite{countg} is a graph $G'$ such that $f(G') = 1 - f(G)$, while being as close as possible to $G$.
Previous work has focused on generating counterfactual explanations for graphs by changing the most elementary unit of a graph, i.e., an existing edge that might be removed, or a non-existing edge that might be added~\cite{countg}. However, as other researchers have observed \cite{perotti2022graphshap}, an explanation language based on the most fundamental unit of a graph structure, might be too fine-grained for producing interesting explanations. Aiming at a higher-order language for producing counterfactual graphs, we turn our attention to some of the main characterizing features of real-world complex networks. In fact, social and technological networks, as well as biological networks (such as brain networks, metabolic and regulatory networks), are all characterized by some common structural features, such as: \emph{network transitivity}, which is the property that two nodes that are both neighbors of the same third node have a high probability of also being connected (a.k.a. \emph{triadic closure}), the existence of \emph{repeated local motifs} and, more importantly, the organization into \emph{communities} or dense modules \cite{milo2002network,pnas.122653799,pnas.0601602103}.
Indeed, the extraction
of dense substructures in networks, such as \emph{maximal cliques}, \emph{quasi-cliques}, $k$-\emph{plex}, $k$-\emph{club}, etc., has received
substantial attention in the algorithmic literature (see~\cite{aggarwal,gionis2015dense,WU2015693,chang2018cohesive,farago2019survey,fang2022cohesive} for surveys). Finding groups of nodes that are densely connected inside and sparsely connected with the outside, is a key concept that has been approached under several different names, including \emph{graph clustering}, \emph{graph partitioning}, \emph{spectral clustering}, and \emph{community detection}~\cite{schaeffer2007graph,fortunato2010community,malliaros2013clustering,nascimento2011spectral,bulucc2016recent}.

Following this observation, in this work, we propose to produce counterfactual graphs based on the alteration of dense substructures. For our purposes, we define a general
\emph{density-based counterfactual search} framework to generate instance-level counterfactual explanations for graph classifiers, which can be instantiated with different notions of dense substructures. This framework identifies the most informative regions of the graphs and manipulates them by adding or removing dense structures until a counterfactual is found. We then instantiate the general framework to specific special cases. In Section \ref{sec:tri}, we present a method (\tri) that, inspired by network transitivity, searches for counterfactual graphs by opening or closing triangles.
Then in Section \ref{sec:cli}, we move to a counterfactual search driven by maximal cliques (\cli).

Our framework can be instantiated with any notion of a dense structure, or region of interest: for instance, in the context of brain networks, ROIs are usually grouped into distinct partitions (\emph{brain parcellation}), according to several properties such as structural and functional markers. Counterfactual graphs generated using the language of density w.r.t. these coarser-grain and well-established taxonomies, might produce explanations that are more consistent with the terminology used to describe the organization of the brain, and thus more comprehensible for the neuroscientists. In fact, deviations in the functional connections among the brain
regions from the normal pattern of connectivity are typically associated
with functional impairments: as a consequence, the notions of \emph{hyper-connectivity} or  \emph{hypo-connectivity} within and between specific regions, are heavily adopted by neuroscientists as fingerprints of specific disorders. Subgraphs that are dense in one class and sparse in the other, have also been proven effective in discriminating between a condition group and
a control group \cite{cslanciano}.

\begin{figure}[t]
\begin{minipage}[b]{\textwidth}
    \centering
    \includegraphics[clip,width=\textwidth]{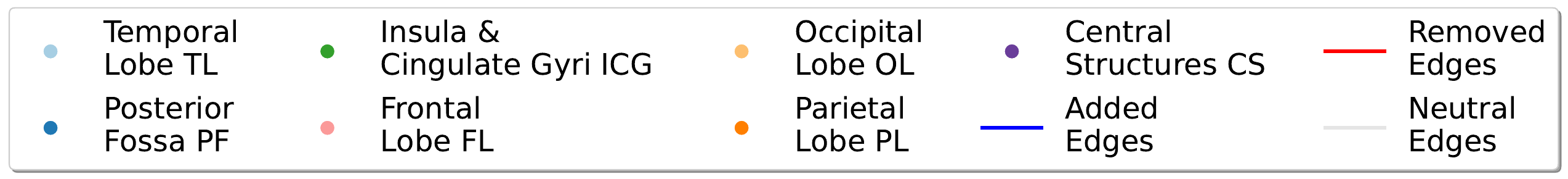}
   \end{minipage}
    \begin{minipage}[b]{1\textwidth}
    \centering
    \includegraphics[width=\textwidth]{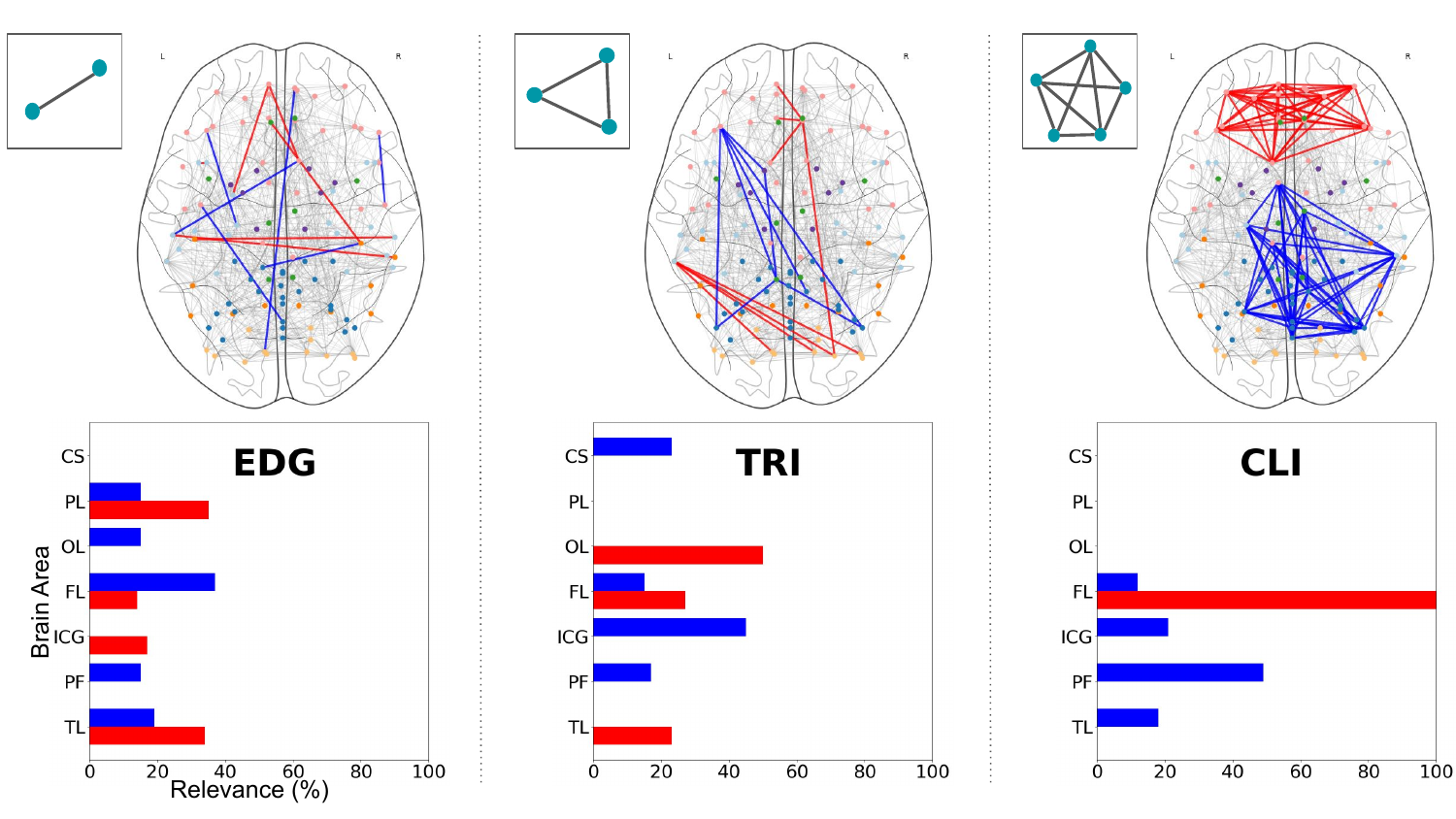}
   \end{minipage}
   \caption{Graph counterfactuals found by \edg (\cite{countg}), \tri, and \cli, for the brain network of patient 9 of the AUT dataset (see Section \ref{sec:results} for details). For each counterfactual, we highlight the edges added and removed in the input network. The boxplots aggregate the changes by brain region for more concise explanations.}
   \label{fig:first_fig}
   \vspace{3mm}
\end{figure}

Figure~\ref{fig:first_fig} showcases an example of our proposal over a brain network from the Autism Brain Image Data Exchange (ABIDE) dataset~\cite{craddock2013neuro} (more details in Section \ref{sec:results}). Nodes in different brain areas are denoted with different colors. In particular, the figure shows three different counterfactual graphs for the same brain network (patient 9): the leftmost one is generated using \cite{countg} (edge-based), the central one is produced using \tri, and the right-most one is created by \cli.
In each counterfactual graph, red edges identify the regions sparsified (\emph{removed edges}), while blue edges indicate the regions densified (\emph{added edges}).
The counterfactual statement corresponding to the counterfactual graph produced by \cli can be expressed in English as follows:
\begin{displayquote}
\textit{Patient X is classified as Autism Spectrum Disorder. If X's brain had less activation in the \textsc{\color{red} Frontal Lobe} and more co-activation between the \textsc{\color{blue} Posterior Fossa}, \textsc{\color{blue} Insula Cingulate Gyri}, and the \textsc{\color{blue} Temporal Lobe} then X would have been classified as Typically Developed.}
\end{displayquote}

\spara{Summary of contributions and roadmap.}
The contributions of this paper can be summarised as follows:
\begin{itemize}
    \item We propose to use the language of dense substructure to guide the search for counterfactual graphs and thus to produce more comprehensible post-hoc counterfactual explanations for graph classifiers.
    \item We define a general and flexible framework, dubbed {\dcs}, that can be instantiated to find counterfactual graphs leveraging different notions of a dense substructure of interest. Our framework is highly modular, providing users with a great deal of flexibility in defining the various parameters involved in the search process. Users can specify how to densify and sparsify (i.e., dense structures of interest), how to rank the nodes to identify the regions to modify, which black-box classifier to use, and whether the search should be refined via perturbation as post-processing.
    \item We showcase in detail two instantiations of {\dcs}: a triangle-based counterfactual search (\tri) and a clique-based counterfactual search (\cli). Both approaches can be further customized, and we present a variation, \rcli, which identifies relevant regions to modify by leveraging the brain's parcellation. This variation can further improve the search process and enhance the interpretability of the explanations.
    \item We evaluate {\dcs} in seven brain networks datasets and compare it with two baseline methods, demonstrating the efficiency of the proposed method and the high interpretability of the explanations it generates.
\end{itemize}

After an overview of the related work (\Cref{sec:sota}), we introduce the density-based counterfactual search problem (\Cref{sec:problem}).
\Cref{sec:dcs} presents our framework to generate counterfactual graphs adhering to the proposed density-oriented language, while \Cref{sec:tri} and \Cref{sec:cli} describe two implementations with customizable parameters.
Finally, \Cref{sec:results} shows our experimental evaluation of the framework and \Cref{sec:conclusions} discusses advantages and limitations.



\section{Related Work}\label{sec:sota}
Post-hoc explanation methods have become essential for understanding the behavior of black-box machine learning models. One such method is counterfactual explanations~\cite{Wachter2017}, which produces example-based explanations by means  of a counterfactual instance for each instance being classified.
More specifically, counterfactual explanations need to exhibit two key characteristics: they must be \emph{similar} to the original instance while being classified in the opposite class of the original instance.
Numerous methods have been proposed to generate counterfactual explanations that possess these critical characteristics~\cite{guidotti2022counterfactual}.

\spara{Explanations for Graph Classifiers.}
There has been a growing interest in addressing the challenge of explaining graph classifiers, resulting in a surge of the number of proposed methods, providing either local or global explanations.
A recent survey \cite{DBLP:journals/corr/abs-2012-15445} categorizes the main (local or) instance-level techniques into four main classes.
\textit{Gradient/feature-based methods}, such as SA and Guided BP ~\cite{baldassarre2019explainability}, and CAM and Grad-CAM~\cite{pope2019explainability}, aim to evaluate the relevance of each feature in the classification task.
\textit{Perturbation-based methods}, such as GNNExplainer~\cite{DBLP:journals/corr/abs-1903-03894}, PGExplainer~\cite{luo2020parameterized}, ZORRO~\cite{funke2020hard}, GraphMask~\cite{schlichtkrull2020interpreting}, RC-Explainer~\cite{wang2020causal}, SubgraphX~\cite{yuan2021explainability}, measure the impact of the perturbation of the input features on the output of the classifier, to detect the most important features.
Among them, GraphShap~\cite{perotti2022graphshap} generates graph-level explanations by ranking a set of input motifs according to their Shapley values.

\textit{Decomposition methods} for graph neural networks, such as LRP~\cite{baldassarre2019explainability}, Excitation BP~\cite{pope2019explainability} and GNN-LRP~\cite{schnake2020higher}, generate feature importance scores by back-propagating decomposed prediction scores to the input layer of the network.
\textit{Surrogate methods}, such as GraphLime~\cite{huang2022graphlime}, RelEx~\cite{zhang2021relex}, and PGM-Explainer~\cite{vu2020pgm}, fit an interpretable model in the neighborhood of the input graph.

Only a few works provide (global or) model-level explanations.
Among them, XGNN~\cite{yuan2020xgnn} is based on graph generation.

\spara{Counterfactual Explanations for Graph Classifiers.}
DBS and OBS~\cite{countg} propose heuristics to locally perturb a generic input graph.
Specifically, they consider two types of modifications: edge addition and edge removal.
The counterfactual explanations are found using a bidirectional search approach that first identifies a feasible counterfactual graph, and then modifies the candidate graph to make it more similar to the input graph.
On the other hand, targeted approaches have been proposed for molecular graphs~\cite{DBLP:journals/corr/abs-2102-03322,wellawatte2022model}, which are graphs where nodes represent atoms and edges are bonds.

CF-GNNExplainer~\cite{DBLP:journals/corr/abs-2102-03322} is a counterfactual version of GNNExplainer~\cite{DBLP:journals/corr/abs-1903-03894} that returns relevant subgraphs as explanations. This method removes edges using a matrix sparsification technique that  minimizes the number of edges changed.
MMACE~\cite{wellawatte2022model} generates counterfactuals for molecular graphs by exploring the chemical space vis the Superfast Traversal, Optimization, Novelty, Exploration and Discovery (STONED) method. In addition, the method uses DBSCAN to generate multiple counterfactuals.

This work proposes a more general framework for counterfactual graph generation that goes beyond existing approaches such as DBS, OBS~\cite{countg} and CF-GNNExplainer~\cite{DBLP:journals/corr/abs-2102-03322}.
While previous works primarily focused on modifying the structure of the original graph by adding or removing one edge at a time, our framework provides more fine-grained control over the graph modifications, as it operates on the dense and sparse regions of the graph.
This opens up possibilities for generating counterfactuals for various scenarios.
\section{Preliminaries}\label{sec:problem}
Given a set of nodes $V$ we denote $\mathcal{G}$ the set of all possible graphs defined over $V$. Given one such graphs $G = (V,E) \in \mathcal{G}$, a \emph{subgraph} of $G$ is a graph $H = (V_H, E_H)$ such that $V_H \subseteq V$ and $E_H \subseteq E$.
A subgraph $H$ is a \emph{k-clique} iff $|V_H| = k$ and $E_H = V_H \times V_H$.
The \emph{density}\footnote{Density is usually defined as the number of edges over the number of possible edges. W.l.o.g. we omit the denominator.} $\delta(H)$ of a subgraph $H$ is defined as its number of edges, i.e., $\delta(H) = |E_H|$.

We assume we are given a binary graph classification model $f: \mathcal{G} \rightarrow \{0,1\}$, that assigns a label in $\{0,1\}$ to each graph in $\mathcal{G}$.
We assume that $f$ \textbf{(i)} is a trained machine learning model whose internal structure is not known (black-box model), \textbf{(ii)} can be queried at will, and \textbf{(iii)} does not change from one query to the other one (i.e., it is static).

Given a specific graph $G \in \mathcal{G}$ a \emph{counterfactual graph} is another graph $G' \in \mathcal{G}$ such that  $f(G) = 1 - f(G')$. Depending on the domain at hand, several desired properties might guide the search for counterfactuals, such as, e.g., \emph{similarity} between the original and the counterfactual instance, \emph{sparsity} (the change affects only a few features), \emph{efficiency} (the search should be fast), and the \emph{feasibility} (to generate a feasible instance).
We will discuss some of these measures in Section \ref{sec:results}.
For the moment, we only need to define the \emph{distance} $\textsf{d}(G, G')$ between two graphs $G$ and $G'$ as the symmetric difference between their edge sets:
\begin{equation}\label{eq:xor}
\textsf{d}(G, G') = |E \setminus E'| + |E' \setminus E|\enspace.
\end{equation}

We next introduce a novel framework to generate counterfactual graphs based on the manipulation of dense substructures, which become the fundamental units of the vocabulary of the explanations produced.



\section{Density-based Counterfactual Search} \label{sec:dcs}
We next introduce our general \emph{Density-based Counterfactual Search} framework (\dcs), which builds instance-level counterfactual explanations by iteratively searching for sparse regions to densify and for dense regions to sparsify.
Pseudocode of {\dcs} is provided in \Cref{alg:db_counter}.

\begin{algorithm}[thb]
\caption{\dcs}
\small
\label{alg:db_counter}
    \begin{algorithmic}[1]
    \Require Binary Graph Classifier $f$; Graph $G$
    \Ensure Counterfactual $G'$
    \State $G' \gets G$
    \While{$f(G) = f(G')$}
        \State \textsc{Densify} a sparse region in $G'$
        \State \textsc{Sparsify} a dense region in $G'$
    \EndWhile
    \State \Return $G'$
    \end{algorithmic}
\end{algorithm}

The algorithm iteratively changes the input graph $G$ until the modified graph $G'$ is classified in the opposite class of $G$.
At each iteration, it adds a dense structure to a sparse region in $G'$ and removes a dense structure from a dense region in $G'$.
Since two different regions of the graph undergo changes at each iteration, $G'$ gradually diverges from the original graph $G$ as the number of iterations increases.
In generating counterfactual graphs, a commonly desired objective is to produce graphs that closely resemble the input graphs. This is because such counterfactual graphs are more likely to provide a concise and interpretable explanation.
For this reason, the algorithm returns the counterfactual graph found in the smallest number of iterations.

\Cref{alg:db_counter} can accommodate any definition of a dense substructure.
In the rest of this section, we introduce two alternative approaches for defining the operations of densification and sparsification.
The first approach, \tri, is based on triadic closure; while the second approach, \cli, is based on maximal cliques.


\subsection{Triangle-based Counterfactual Search}\label{sec:tri}
The \emph{Triangle-based Counterfactual Search} ({\tri}) is illustrated in \Cref{alg:TSearch}.
In addition to the original graph $G$ and the classifier $f$, {\tri} takes as input a sorted list of candidate edges to remove $E_-$ (to destroy triangles) and a sorted list of candidate edges to add $E_+$ (to create triangles) in the counterfactual graph. These lists are prepared using \Cref{alg:TScore} (discussed below).
\tri iterates over the two lists until the graph becomes a counterfactual graph for $G$.
At each iteration $i$, it selects the next best edge to add ($e_+$) and to remove ($e_-$) from the current graph $G_i$.
If all the possible wedges have been closed, or if all the possible triangles have been opened (i.e. there are no more edges available in either $E_{-}$ or $E_{+}$), but $G_i$ is still classified in the same class as $G$, the algorithm returns $\emptyset$, indicating that a counterfactual could not be found.

\begin{algorithm}[thb]
    \small
    \caption{\tri}
    \label{alg:TSearch}
    \begin{algorithmic}[1]
    \Require Binary Graph Classifier $f$; Graph $G$
    \Require Sorted lists of candidate edges to remove $E_{-}$ and to add $E_{+}$
    \Ensure Counterfactual $G'$ if found; $\emptyset$ otherwise
    \State $G_0 \gets G$; $i \gets 1$
    \While{$f(G_{i-1}) = f(G)$ \textbf{and} $i \leq \min\left(|E_{-}|, |E_{+}|\right)$}
        \State $e_{-} \gets \Call{nextBest}{E_{-}}$; $e_{+} \gets \Call{nextBest}{E_{+}}$
        \State $E_i \gets E_{i-1} \setminus \{e_{-}\} \cup \{e_{+}\}$; $i \gets i + 1$
    \EndWhile
    \If{$f(G_i) \neq f(G)$} \Return $G_i$
    \Else{ \Return $\emptyset$ }
    \EndIf
    \end{algorithmic}
\end{algorithm}

Given a graph $G$, \Cref{alg:TScore} first computes a score for each feasible edge $(u,v) \in V\times V$, and then partitions the edges into two lists: a list of existing edges that could be removed ($E_{-}$) and a list of non-existing edges that could be added ($E_{+}$).
Finally, the algorithm sorts both lists based on the number of triangles in $G$ that contain the vertices of each edge.
In particular, $E_{-}$ is sorted in ascending order, while $E_{+}$ is sorted in descending order.
This sorting strategy ensures that {\tri} adds triangles in sparse regions of $G$, and removes triangles from the dense regions.

\begin{algorithm}[thb]
    \small
    \caption{\textsc{Triangle Score}}
    \label{alg:TScore}
    \begin{algorithmic}[1]
    \Require Graph $G$
    \Ensure Sorted lists of candidate edges to remove $E_{-}$ and to add $E_{+}$
    \State $\mathsf{T}(v) \gets$ number of triangles including $v$, $\forall v \in V$
    \State $E_{-} \gets E_{+} \gets \emptyset$
    \For{$(v,u) \in V\times V$}
        \State $s_v \gets \mathrm{T}(v)$; $s_u \gets \mathrm{T}(u)$
	\If{$(v,u) \in E$}
		$E_{-} \gets E_{-} \cup \{(s_v+s_u,(v,u))\}$
        \Else
            { $E_{+} \gets E_{+} \cup \{(s_v+s_u,(v,u))\}$ }
	\EndIf
    \EndFor
    \State $Sort_s (E_{-})$ in ascending order of score;
    \State $Sort_s (E_{+})$ in descending order of score;
    \State \Return $(E_{-}, E_{+})$
    \end{algorithmic}
\end{algorithm}

\subsection{Clique-based Counterfactual Search}\label{sec:cli}
The \emph{Clique-based Counterfactual Search} ({\cli}), illustrated in \Cref{alg:DS}, follows a structure similar to \Cref{alg:db_counter} but employs several heuristics to speed up the search for a counterfactual graph $G'$ for $G$.
The algorithm receives in input two additional parameters: the maximum number of iterations $\textsf{max}_I$, and the list of nodes in $G$ ranked according to a metric that gives more importance to nodes that belong to dense regions in $G$.
At each iteration $i$, the algorithm adds a clique to a sparse region of the current graph $G_i$ around the next worst node in the ranking $\bar{n}_h$ and removes a maximal clique from a dense region in $G_i$ around the next best node in the ranking $\bar{n}_l$.
The algorithm terminates when either $G_i$ is classified in the opposite class of $G$ or the maximum number of iterations $\textsf{max}_I$ is reached.
The densification of a sparse region is carried out by \Cref{alg:HS} (\textsc{DensifyCLI}) and the sparsification by \Cref{alg:LS} (\textsc{SparsifyCLI}).
In the following, a clique is represented by its set of vertices, as its set of edges is the set of all the possible edges between such nodes.

\begin{algorithm}[thb]
    \small
    \caption{\cli}
    \label{alg:DS}
    \begin{algorithmic}[1]
    \Require Binary Graph Classifier $f$; Graph $G$
    \Require Max Num Iteration $\textsf{max}_I$; Sorted list of vertices $\bar{V}$
    \Ensure Counterfactual $G'$ if found; $\emptyset$ otherwise
    \State $D[v] \gets 0$, $\forall v \in V$
    \State $G_0 \gets G$; $\mathcal{L} \gets \emptyset$; $i \gets 1$
    \While{$f(G_{i-1}) = f(G)$ \textbf{and} $i \leq \textsf{max}_I$}
        \State $\bar{n}_l,\bar{n}_h \gets \textsc{nextBest}(\bar{V}), \textsc{nextWorst}(\bar{V})$
	\State $G_i \gets \Call{SparsifyCLI}{G, G_{i-1}, \bar{n}_l, D, \mathcal{L}}$
	\State $d_h\gets 0$; $s \gets d_l \gets \textsf{d}(G_{i-1}, G_{i})$\label{line:xor}
	\While{$f(G_i) = f(G)$ \textbf{and} $d_l > d_h$}
            \State $s \gets s - d_h$
		\State $G^h \gets \Call{DensifyCLI}{G_{i},\bar{n}_h,D,s}$
		\State $d_h \gets \textsf{d}(G_{i}, G^h)$; $G_{i} \gets G^h$\label{line:xor2}
	\EndWhile
	\State $i \gets i + 1$		
    \EndWhile
    \If{$f(G_i) \neq f(G)$} \Return $G_i$
    \Else{ \Return $\emptyset$ }
    \EndIf
    \end{algorithmic}
\end{algorithm}

Procedure~\textsc{SparsifyCLI} identifies a maximal clique in the input graph $G$ surrounding a given node $n$, and removes all the edges in that clique in the candidate counterfactual graph $G'$.
The algorithm operates by identifying the largest clique in $G$ including $n$ that has the lowest overlap with the cliques removed in prior iterations.
By choosing the largest, lowest-overlap clique, the algorithm sparsifies one of the densest regions in $G'$.
Note that the cliques considered by the algorithm are found in the original graph $G$, and some of their edges may have already been removed from $G'$ in previous iterations of \textsc{SparsifyCLI}.
After the desired clique $C$ has been identified, the algorithm removes all its edges from $G'$ and stores $C$ in the set of cliques removed $\mathcal{L}$. Finally, the counts associated with each node in the clique are incremented by 1.

\begin{algorithm}[thb]
    \small
    \caption{\textsc{SparsifyCLI}}
    \label{alg:LS}
    \begin{algorithmic}[1]
    \Require Graph $G$; Candidate Counterfactual $G'$
    \Require Node $n$; Dictionary $D$; Set of Cliques $\mathcal{L}$
    \Ensure $G'$ with a clique removed
    \State $\mathcal{C} \gets$ cliques in $G$ including $n$
    \If{$\mathcal{L} = \emptyset$}
	\Return largest clique in  $\mathcal{C}$
    \EndIf
    \State $O \gets \emptyset$
    \For{$C \in \mathcal{C}$}
	\State $o \gets \max_{L \in \mathcal{L}}{(|C \cap L|)}$; $O \gets O \cup \{(o, C)\}$
    \EndFor
    \State $C \gets $ clique in $O$ with lowest value $o$
    \State $\mathcal{L} \gets \mathcal{L} \cup \{C\}$
    \State $D[v] \gets D[v] + 1$ for each $v$ in $C$\label{line:dict}
    \State \Return $G' \setminus \{$ edges in $C \,\}$
    \end{algorithmic}
\end{algorithm}

\begin{algorithm}[thb]
    \caption{\textsc{DensifyCLI}}
    \label{alg:HS}
    \begin{algorithmic}[1]
    \Require Graph $G$; Node $n$; Dictionary $D$; Size $s$
    \Ensure $G$ with a clique added
    \State $\Gamma_G(n) \gets$ 2-hop neighborhood of $n$ in $G$ sorted according to $D$ (ascending)\label{line:2hop}
    \State $W \gets V \setminus \Gamma_G(n)$ sorted according to $D$ (ascending)\label{line:others}
    \State $C \gets \Call{concat}{\Gamma_G(n), W}$; $V_s \gets $ first $s$ nodes in $C$
    \State $D[v] \gets D[v] - 1$ for each $v \in V_s$
    \State \Return $G \cup \{$ edges between the nodes in $V_s \,\}$
    \end{algorithmic}
\end{algorithm}

Procedure~\textsc{DensifyCLI} is iteratively called until the dense region added to the candidate counterfactual graph has at least as many edges as the dense region removed by Algorithm~\ref{alg:LS}~\footnote{In our experiments we constrained the max deviation between the number of edges added and removed in terms of the max number of nodes $b$ that a clique added can have with respect to the number of nodes in the clique removed, and set $b = 10$.}.
This ensures that the size of the counterfactual is similar to that of the original graph.
At each iteration, the algorithm identifies a sparse region of size $s$ around the given node $n$ and adds all the possible edges between the $s$ nodes.
The size of the region $s$ is determined by subtracting the number of edges added in the previous iterations from the number of edges removed by Algorithm~\ref{alg:LS}.
To avoid densifying a region that has just been sparsified, the algorithm selects a sparse region involving nodes that are not present in many cliques added in previous iterations.
To achieve this, the algorithm uses a dictionary $D$, which keeps track of the number of times each node has been part of a clique added to the candidate counterfactual.
Given a node $n$, \textsc{DensifyCLI} sorts both the 2-hop neighborhood $\Gamma_G(n)$ of $n$ and the rest of the vertices $W$ according to their counts in $D$ (Algorithm~\ref{alg:HS} lines~\ref{line:2hop}-\ref{line:others}).
Then, it adds to $G$ the clique $C$ consisting of the first $s$ nodes in the concatenation between $\Gamma_G(n)$ and $W$, and updates $D$ by decreasing the counts associated with the nodes in $C$.

\spara{Further customizability of the framework.}
In \Cref{alg:DS}, we made specific design choices that, however, can be customized and adapted to meet the needs of the application at hand.

Firstly, \Cref{alg:DS} receives as input the list $\bar{V}$  of nodes in $G$ ranked according to a metric that prioritizes nodes in dense regions for \Cref{alg:LS} and nodes in sparse regions for \Cref{alg:HS}. In our implementation of \cli\ used in the experiments in Section \ref{sec:results} we sort the nodes in $\bar{V}$ based on the number of triangles in which each node participates.
However, alternative measures could be used to rank the nodes, such as the node clustering coefficient or other features at node level.
The key is to select a ranking that allows for traversing in one direction to be a good heuristic for sparsifying regions while traversing in the other direction is a good heuristic for densifying regions.

When domain-specific information is available, it is important to customize the algorithm to take such information into consideration. In the case of brain networks, for instance, 
nodes can be partitioned into well-defined and distinct regions (i.e. the brain lobes).
In \Cref{sec:results}, we explore a variation of \cli that selects the regions to sparsify/densify based on the brain lobes, which we refer to as \rcli.
The {\rcli} algorithm uses a two-level ranking strategy that first ranks the brain lobes according to the density of the subgraph induced by their nodes (regions having higher density ranked higher) and then, within each region, ranks the nodes according to the number of triangles in which they participate.
This two-level ranking allows \rcli to conduct the edge changes within a lower number of brain regions and thus generate more interpretable explanations.


Secondly, for the sake of \emph{feasibility} of the counterfactual, i.e., keeping its density similar to the original graph, we alternate between 
\Cref{alg:LS} and \Cref{alg:HS}, meaning that we call them the same number of times.
However, while this strategy is effective in most cases, it may not be the optimal approach.
An alternative approach might let the density of the original graph govern the calls to \Cref{alg:LS} and \Cref{alg:HS}, so that when the graph is very sparse, \Cref{alg:HS} is called more often than \Cref{alg:LS}, and the other way around.
\section{Experimental Evaluation} \label{sec:results}
We next showcase the application of density-based counterfactuals in the context of brain networks, highlighting the high interpretability of such counterfactuals.

\subsection{Brain Networks}
Brain networks can be constructed using non-invasive techniques such as Functional Magnetic Resonance Imaging (fMRI) in resting-state patients.
By measuring blood flow, fMRI exploits the link between neural activity and blood flow and oxygenation, to associate a time series of activation scores at voxel level.
The voxels' signals are parcellated into Regions of Interest (ROIs) (\textit{nodes of the graph}) using specific templates, such as the Automated Anatomical Labeling (AAL)~\cite{tzourio2002automated} or the 200~\cite{craddock2012whole} parcellation scheme.
Then, interactions between ROIs (\textit{edges of the graph}) are identified by looking at the correlation between the corresponding time series.
Finally, relevant interactions are selected by applying a threshold (edge pruning), to obtain the brain's functional connectome.
ROIs can be further aggregated into areas associated with the lobes of the brain. As discussed before, this aggregation can be exploited to express interpretable density-based counterfactual explanations.

We consider seven publicly available brain network datasets.

\spara{AUT} is a dataset gathered within the Autism Brain Image Data Exchange (ABIDE) \cite{craddock2013neuro} project. This dataset includes brain network data from 49 patients with Autism Spectrum Disorder (ASD, condition group) and 52 Typically Developed (TD, control group) patients, all under the age of 9 years old.

\spara{BIP} dataset about lithium response in type I bipolar disorder patients~\cite{sani2018association}.

\spara{ADHD, ADHDM} come from the Multimodal Treatment of Attention Deficit Hyperactivity Disorder project\footnote{\url{http://fcon\_1000.projects.nitrc.org/indi/ACPI/html/acpi\_mta\_1.html}}, which investigated the impact of cannabis use on adults with or without a childhood diagnosis of ADHD. In the ADHD dataset, subjects are labeled as either ``ADHD'' or ``TD'', while in the ADHDM dataset, they are labeled as ``Marijuana use'' or ``Marijuana not used''.
  
\spara{OHSU, PEK, KKI}\footnote{\url{https://github.com/GRAND-Lab/graph_datasets}} are datasets constructed for three brain classification tasks: Attention Deficit Hyperactivity Disorder classification (OHSU), Hyperactive Impulsive classification (PEK), and gender classification (KKI)~\cite{pan2016task}.

\smallskip

Data is preprocessed following the literature for converting time series to correlation matrices \footnote{See \url{http://preprocessed-connectomes-project.org/abide/dparsf.html} for AUT and BIP, and \url{https://ccraddock.github.io/cluster_roi/atlases.html} for ADHD and ADHDM. For OHSU, PEK, and KKI, the data was already preprocessed.}.
To generate the graph dataset, correlation matrices are transformed into adjacency matrices by setting edges when the correlation between the two nodes is higher than a fixed threshold.
The threshold is selected based on the distribution of the correlation matrix values, using the 90th percentile for ADHD and AUT, and 80th for BIP. All the preprocessed graph datasets are available in our repository\footnote{\url{https://github.com/carlo-abrate/Counterfactual-Explanations-for-Graph-Classification-Through-the-Lenses-of-Density.git}}.

\Cref{tab-data} reports, for each dataset, the number of networks, the percentage of networks in class 1 (since we deal with binary classification, we report values for one class only), the total number of vertices and edges in the networks, and the accuracy and F1 score of the binary classifier trained on the dataset (see below).

\begin{table}[t]
    \centering
    \caption{Num. of graphs $|\mathcal{G}|$, percentage of graphs in class $1$, num. of nodes $|V|$, average num. of edges per graph $\mathrm{avg}|E|$, and accuracy $\textsf{ACC}$ and $\textsf{F1}$ score of the binary classifier, for each dataset.}
	\label{tab-data}
	\begin{tabular}{ccccccc}
	\toprule
	\textbf{Dataset} & \textbf{$|\mathcal{G}|$} & \textbf{$|Y_{=1}|$}  & \textbf{$|V|$} & \textbf{$\mathrm{avg}|E|$} & \textsf{ACC} & \textsf{F1} \\
	\midrule
	AUT & 101 & $48 \%$ & 116 & 665 & 0.92 & 0.90 \\
        BIP & 118 & $46 \%$ & 116 & 667 & 0.66 & 0.54 \\
	ADHD & 123 & $32 \%$ & 116 & 667 & 0.80 & 0.59 \\
	ADHDM & 123 & $50 \%$ & 116 & 667 & 0.93 & 0.93 \\
	OSHU & 79 & $56 \%$ & 190 & 199 & 0.68 & 0.72 \\
	PEK & 85 & $42 \%$ & 190 & 77 & 0.71 & 0.58 \\
	KKI & 83 & $55 \%$ & 190 & 48 & 0.66 & 0.68 \\
	\bottomrule
    \end{tabular}
\vspace{3mm}
\end{table}

\mpara{Classifier.}
The proposed framework is model-agnostic, making it suitable for explaining any kind of binary classifier.
In our experiments, we consider a binary classifier designed for graph classification that exploits the \emph{Spectral Features} (SF)~\cite{sfclassifier} of the graph to determine class memberships.
Let $A \in \{0,1\}^{|V|\times |V|}$ be the adjacency matrix of the graph, $D$ be the diagonal matrix of node degrees, and $L = I - D^{-1/2} A D^{-1/2}$ be the normalized Laplacian of $A$.
The SF of the graph is a vector consisting of the $k$ smallest positive eigenvalues of $L$, sorted in ascending order.
We trained a KNN classifier with various parameter settings and selected the optimal configuration based on the accuracy (\textsf{ACC}) and \textsf{F1} score using 5-fold cross-validation. The values of \textsf{ACC} and \textsf{F1} of the configurations selected are reported in the last two columns of \Cref{tab-data}.

\subsection{Metrics}\label{sec:metrics}
Various metrics have been proposed to evaluate the quality of counterfactual explanations \cite{guidotti2022counterfactual}. The selection of which measures to prioritize over others depends on factors such as the data type, the black-box models considered, and the vocabulary used to formulate the counterfactual statements.
We consider three measures specifically proposed to evaluate graph counterfactuals~\cite{prado2022survey}.

\spara{Flip rate:}
measures the percentage of graphs in the dataset for which the algorithm was able to find a counterfactual explanation~\cite{mothilal2020explaining,prado2022survey}.

\spara{Edit distance:} 
measures how \emph{different} is a graph $G$ from its counterfactual $G'$, and, in our case, is defined as the ratio between the symmetric difference of the edge sets of $G$ and $G'$ (\Cref{eq:xor}) and $|E \cup E'|$:
\[
d_{\%}(G,G') = \frac{\mathsf{d}(G,G')}{|E \cup E'|}\,.
\]
\spara{Calls:}
run-time complexity of a counterfactual search method measured in terms of the number of calls to the black-box model ($C$).

\subsection{Baselines}
We compare the performance of \tri, \cli, and \rcli against three baseline methods. The first baseline, \edg~\cite{countg}, utilizes an edge-based language to generate counterfactual explanations. The second baseline, \dataset, is an instance-level counterfactual search method proposed in \cite{guidotti2022counterfactual}. This method searches for the closest graph in the dataset that is classified by the black-box model in the opposite class and returns it as a counterfactual explanation for the input graph. 

Following~\cite{countg} we also equip the \rcli and \dataset methods with a \emph{backward search} phase which tries to refine the counterfactual found by 
modifying the edges in the symmetric difference between the edge set of the input graph and that of the counterfactual graph, with the aim of reducing the distance between the two graphs. The resulting methods are named \rclibw and \datasetbw respectively.

All the methods are implemented in Python and the code is made publicly available\footnote{\url{https://github.com/carlo-abrate/Counterfactual-Explanations-for-Graph-Classification-Through-the-Lenses-of-Density}} together with the datasets used in our analysis, and a supplemental material document containing further experimental results.
\subsection{Qualitative analysis}\label{sec:local}

\begin{figure}[t!]
\centering
    \begin{minipage}[b]{\textwidth}
    
    \centering
    \includegraphics[clip,width=.9\textwidth]{Figures/brains/legend.pdf}
   \end{minipage}
    \begin{minipage}[b]{0.49\textwidth}
    \centering
    \includegraphics[width=\textwidth]{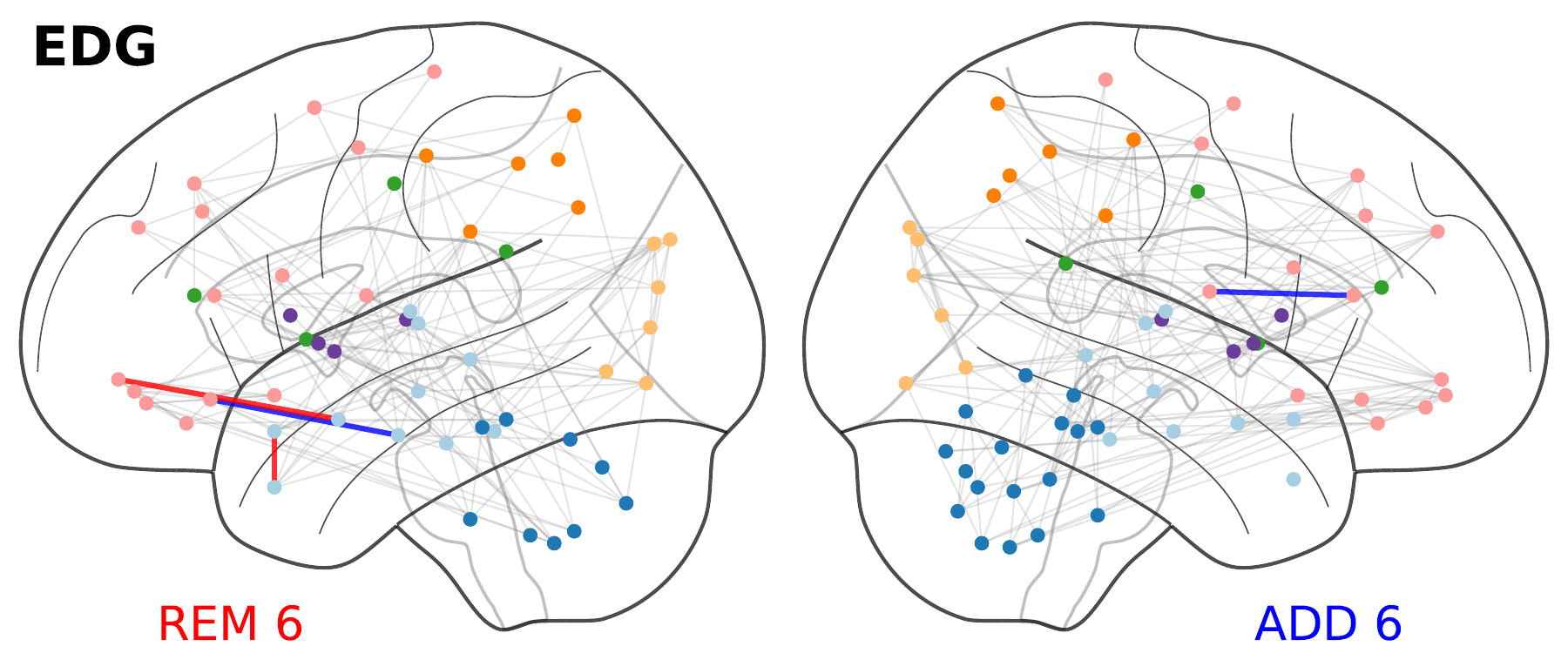}
   \end{minipage}
   \hfill
   \begin{minipage}[b]{0.49\textwidth}
    \centering
    \includegraphics[width=\textwidth]{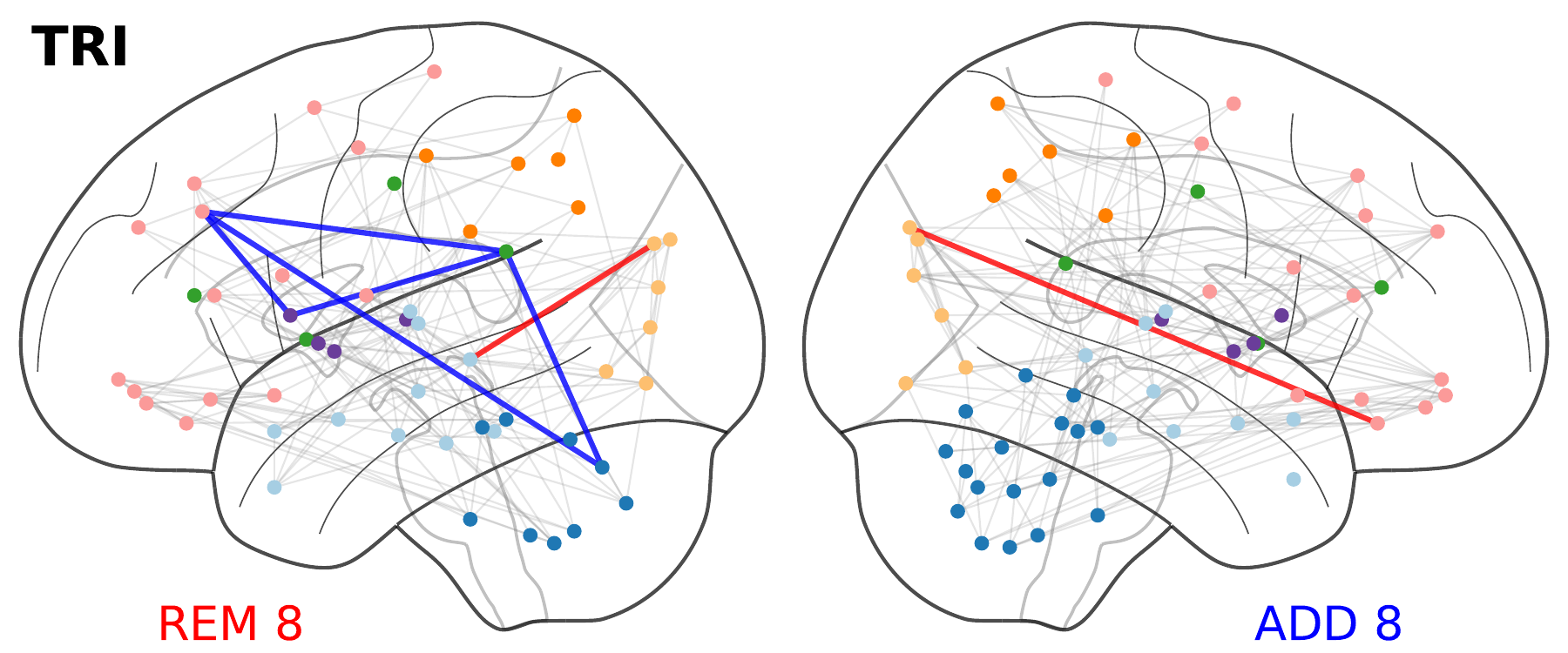}
   \end{minipage}
   \begin{minipage}[b]{0.49\textwidth}
    \centering
    \includegraphics[width=\textwidth]{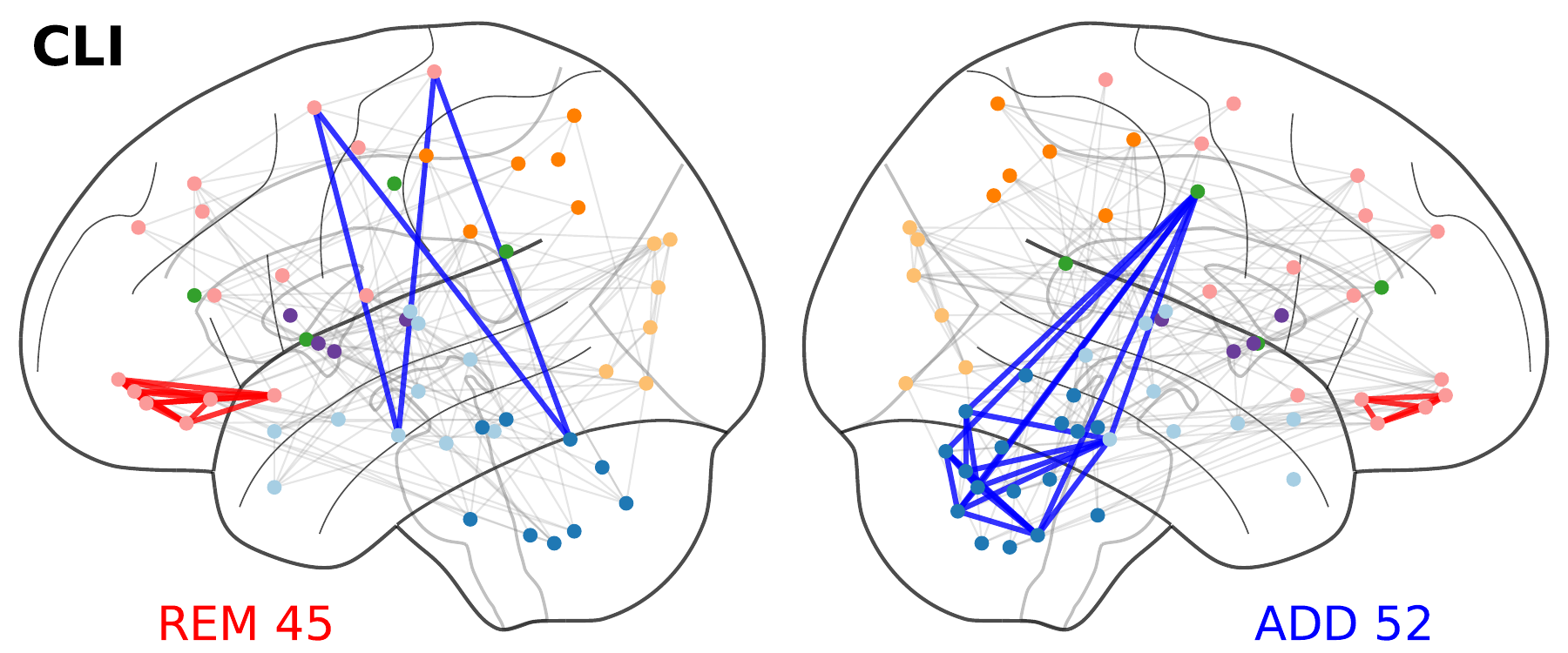}
   \end{minipage}
   \hfill
   \begin{minipage}[b]{0.49\textwidth}
    \centering
    \includegraphics[width=\textwidth]{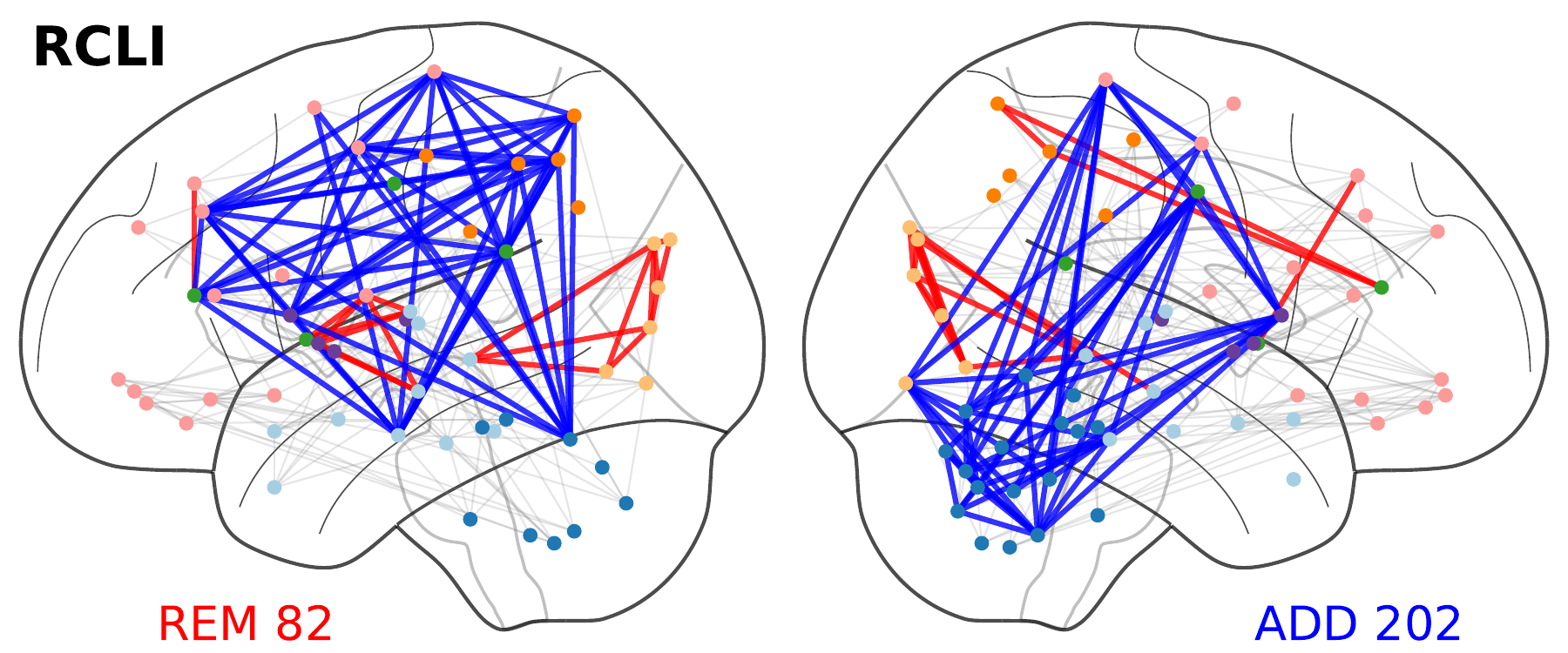}
   \end{minipage}
   \begin{minipage}[b]{0.49\textwidth}
    \centering
    \includegraphics[width=\textwidth]{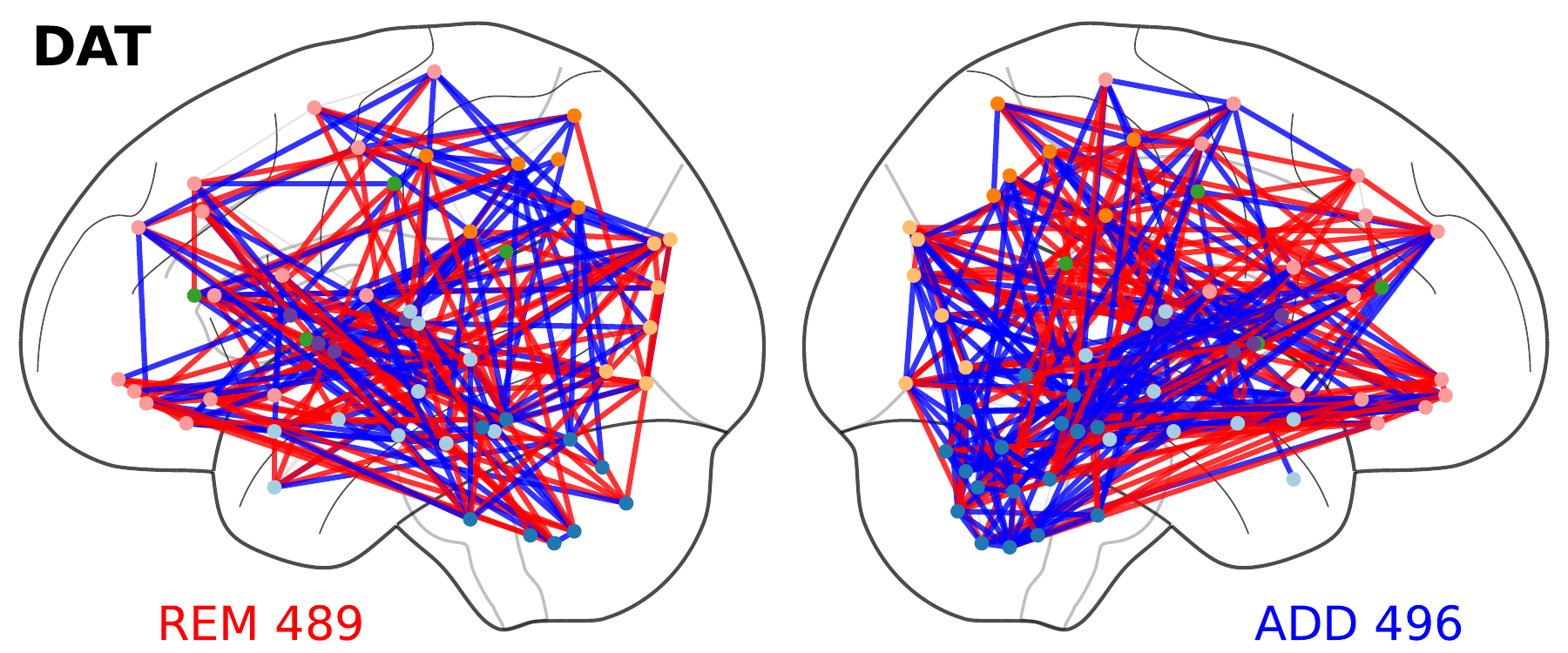}
   \end{minipage}
   \hfill
   \begin{minipage}[b]{0.49\textwidth}
    \centering
    \includegraphics[width=\textwidth]{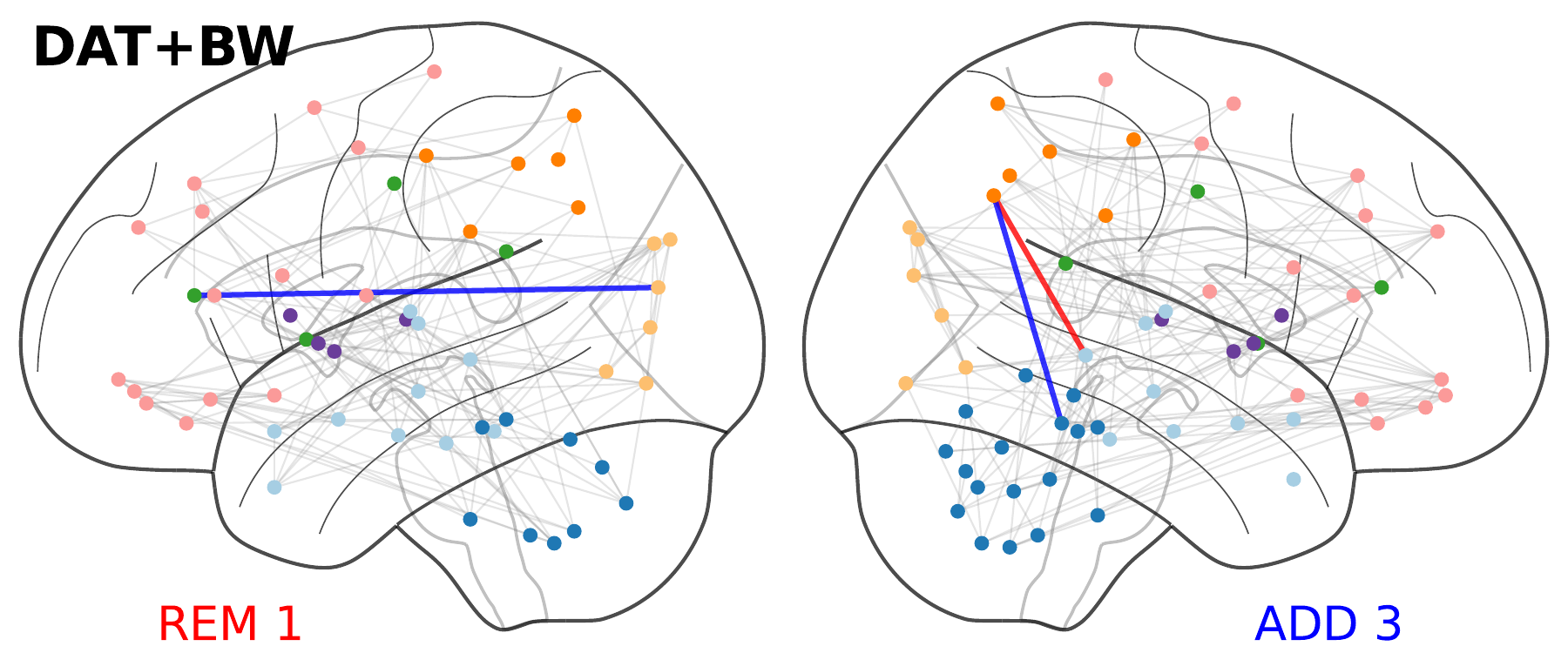}
   \end{minipage}
   \begin{minipage}[b]{0.32\textwidth}
    \centering
    \includegraphics[width=0.8\textwidth]{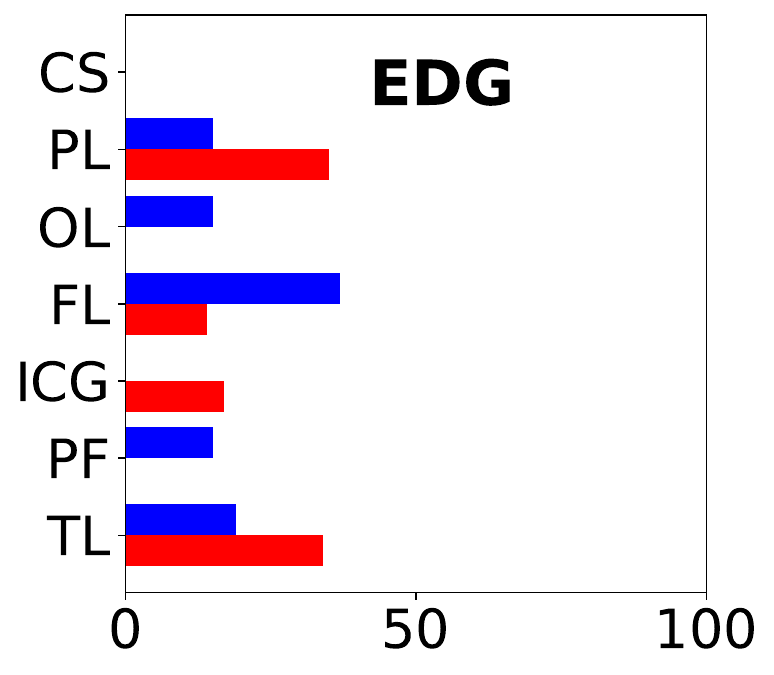}
   \end{minipage}
   \hfill
   \begin{minipage}[b]{0.32\textwidth}
    \centering
    \includegraphics[width=0.8\textwidth]{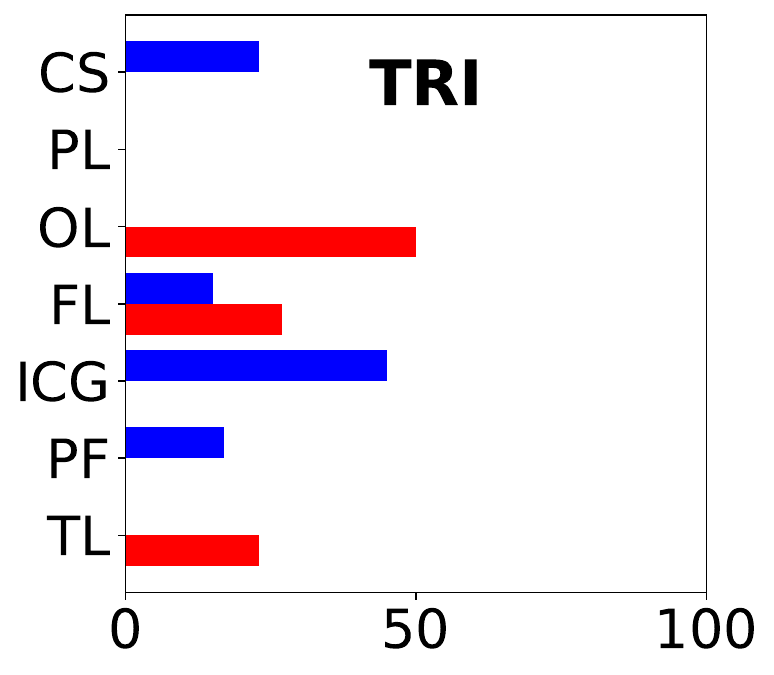}
   \end{minipage}
   \hfill
   \begin{minipage}[b]{0.32\textwidth}
    \centering
    \includegraphics[width=0.8\textwidth]{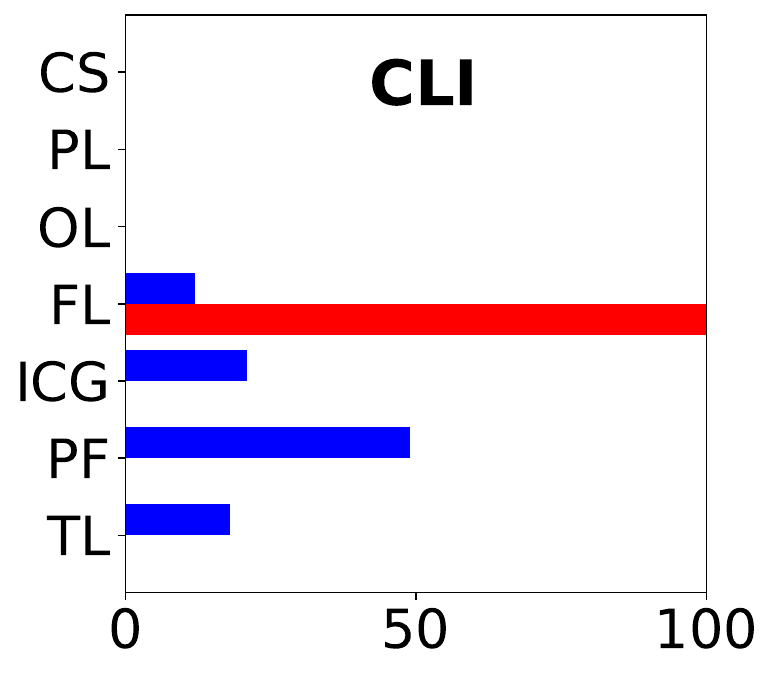}
   \end{minipage}
   \begin{minipage}[b]{0.32\textwidth}
    \centering
    \includegraphics[width=0.8\textwidth]{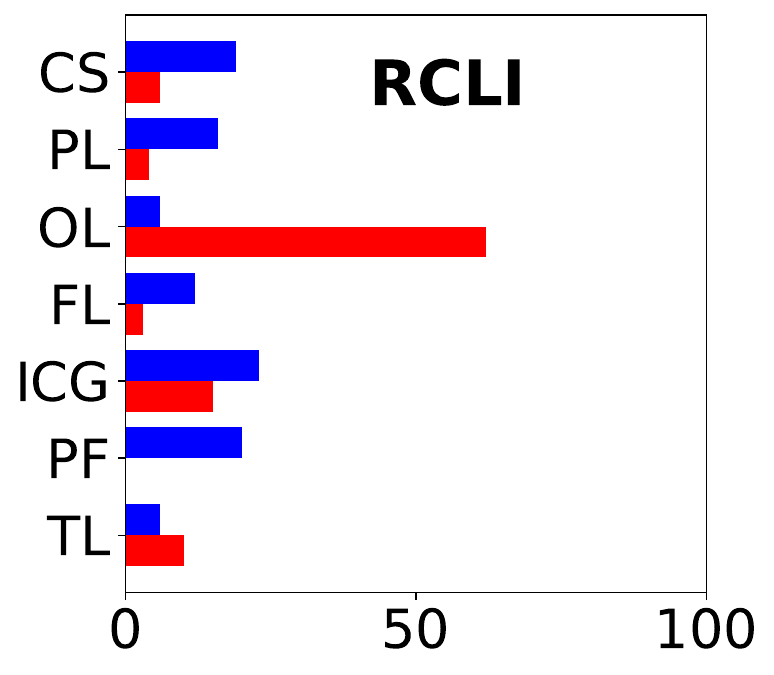}
   \end{minipage}
   \hfill
   \begin{minipage}[b]{0.32\textwidth}
    \centering
    \includegraphics[width=0.8\textwidth]{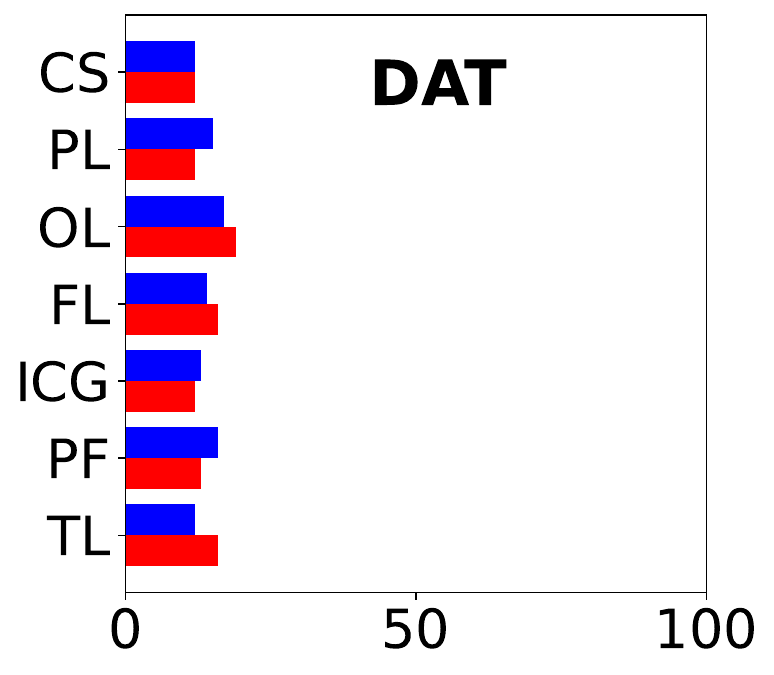}
   \end{minipage}
   \hfill
   \begin{minipage}[b]{0.32\textwidth}
    \centering
    \includegraphics[width=0.8\textwidth]{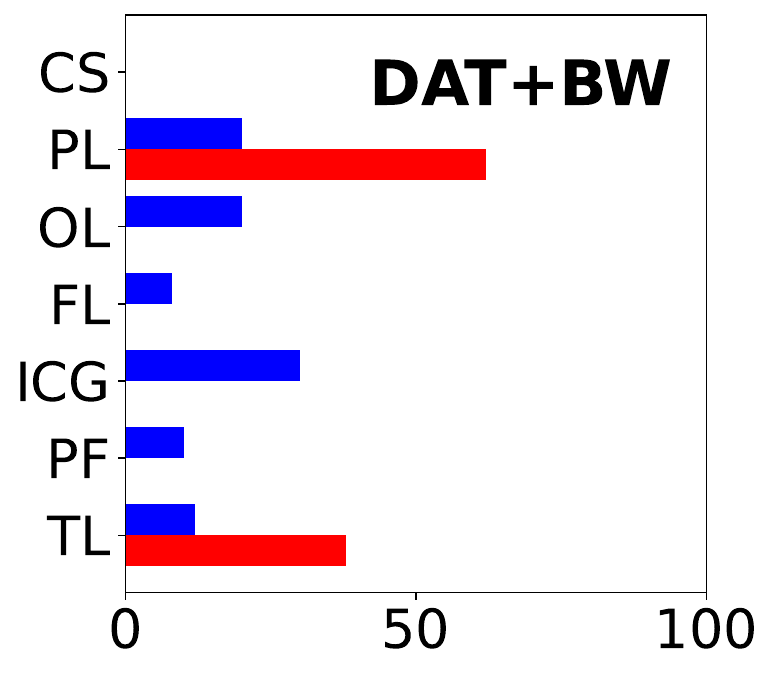}
   \end{minipage}
   \caption{Counterfactual graphs found by each method for patient $9$ in AUT. The upper illustrations show the connectome of the patient where red edges indicate \textit{\color{red}connections to remove}, and blue edges indicate \textit{\color{blue}connections to add} to generate the counterfactual (REM and ADD are the total numbers). The lower barplots show to which brain lobes the ROIs of the edges added (in blue) and removed (in red) belong (as percentages).}
   \label{fig:graph_counterfactual_methods}
\end{figure}

In this section, we compare the counterfactual graphs generated by three instantiations of \dcs, namely \tri, \cli, and \rcli, with those produced by the three baseline methods, \edg, \dataset, and \datasetbw, for specific patients in three datasets. All the results presented pertain to brain networks for which the classifier accurately predicted the class.

\mpara{AUT Dataset.}
\Cref{fig:graph_counterfactual_methods} shows the counterfactual graphs for patient $9$ in AUT. This patient is classified as ``Autism Spectrum Disorder''.
For each method, the left figure shows the connectome of the patient overlaid on the brain glass schematics, where ROIs are projected onto a 2D space of the image, and different colors represent ROIs in different brain lobes. Blue edges indicate edges added to the counterfactual graph, while red edges denote edges removed from the counterfactual graph.
In addition to the connectome visualization, the right barplots illustrate the distribution of changes among brain lobes.
For each brain lobe, the bars report the percentage of the nodes involved in the added (blue) and removed (red) edges that belong to that lobe. By examining these barplots, we can gain insights into which brain regions are most affected by each method's counterfactual graph generation process.
We first observe that each method perturbed different regions of the brain.
This is due to the fact that the set of changes identified by each method depends on a range of factors, including the method's underlying assumptions, its optimization criteria, and its specific implementation.
The choice of counterfactual generation method should take into account the specific properties of the input data and the desired goals of the counterfactual analysis.
One of the desiderata is interpretability.
In general, the larger the number of regions changed and the more homogeneously the changes are distributed within the regions, the less human-interpretable the counterfactual explanation becomes.
Of the methods examined, \dataset produced the most complex explanation, with almost the same number of edges added and removed from each brain lobe.
The \datasetbw method provides a partial solution to this issue by removing edges mainly from the Parietal Lobe and the Temporal Lobe, which reduces the heterogeneity of edge removals across the brain lobes. However, the edge additions still span across many regions, limiting the interpretability of the solution.
The \edg method suffers from similar limitations in that its counterfactual graph involves changes spanning across most of the brain lobes.
In contrast, \tri and \cli provide simpler explanations, as they perturbed a lower number of regions and concentrated most of the changes in the same regions.
Specifically, \tri mainly sparsified the Occipital Lobe and densified the Insula \& Cingulate Gyri, while \cli sparsified only the Frontal Lobe and added most of the edges in the Posterior Fossa.
This results in a more focused and interpretable explanation.
In fact, the output of \cli can be summarized by the following simple \textit{counterfactual statement}:
\begin{displayquote}
\emph{Patient X is classified as Autism Spectrum Disorder. If X's brain had less activation in the \textsc{\color{red} Frontal Lobe} and more co-activation in the \textsc{\color{blue} Posterior Fossa}, \textsc{\color{blue} Insula Cingulate Gyri}, and the \textsc{\color{blue} Temporal Lobe} then X would have been classified as Typically Developed.}
\end{displayquote}

\begin{figure}[t!]
\centering
    \begin{minipage}[b]{\textwidth}
    \centering
    \includegraphics[clip,width=.9\textwidth]{Figures/brains/legend.pdf}
   \end{minipage}
    \begin{minipage}[b]{0.49\textwidth}
    \centering
    \includegraphics[width=\textwidth]{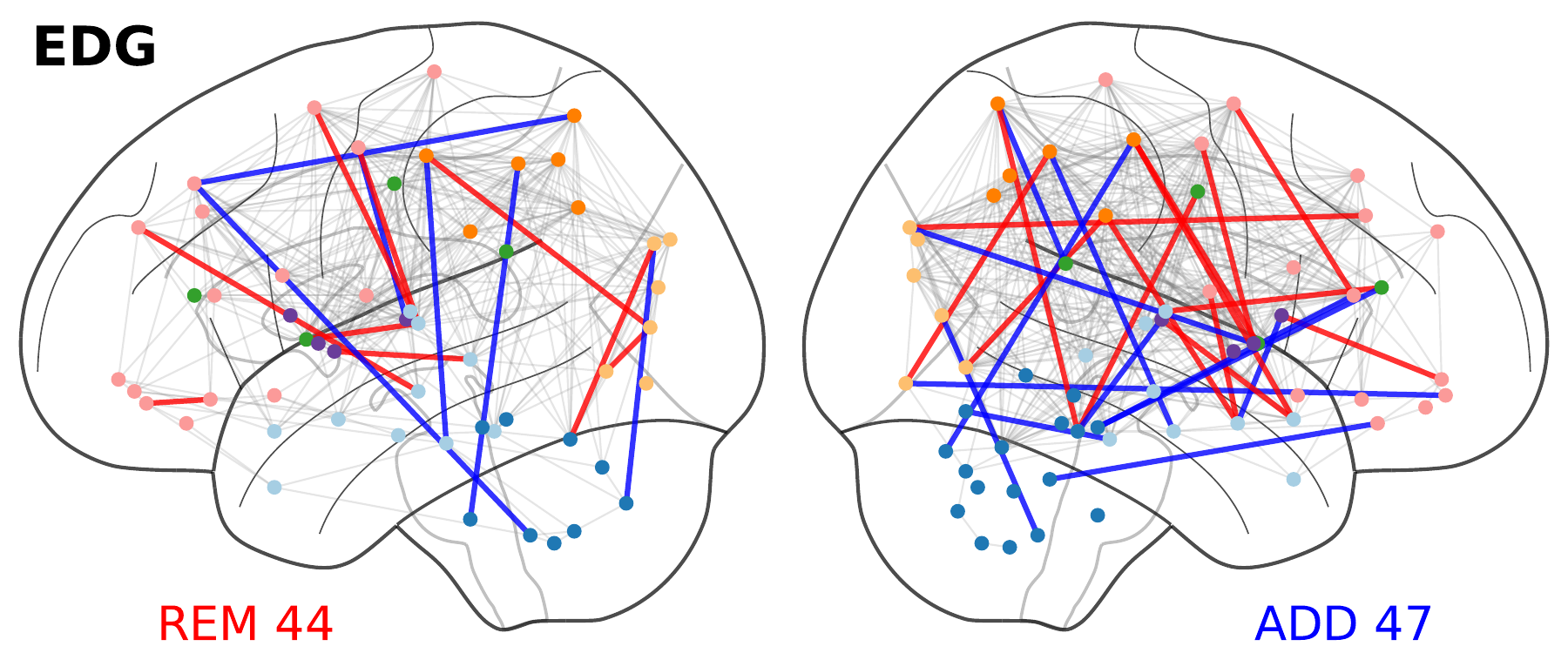}
   \end{minipage}
   \hfill
   \begin{minipage}[b]{0.49\textwidth}
    \centering
    \includegraphics[width=\textwidth]{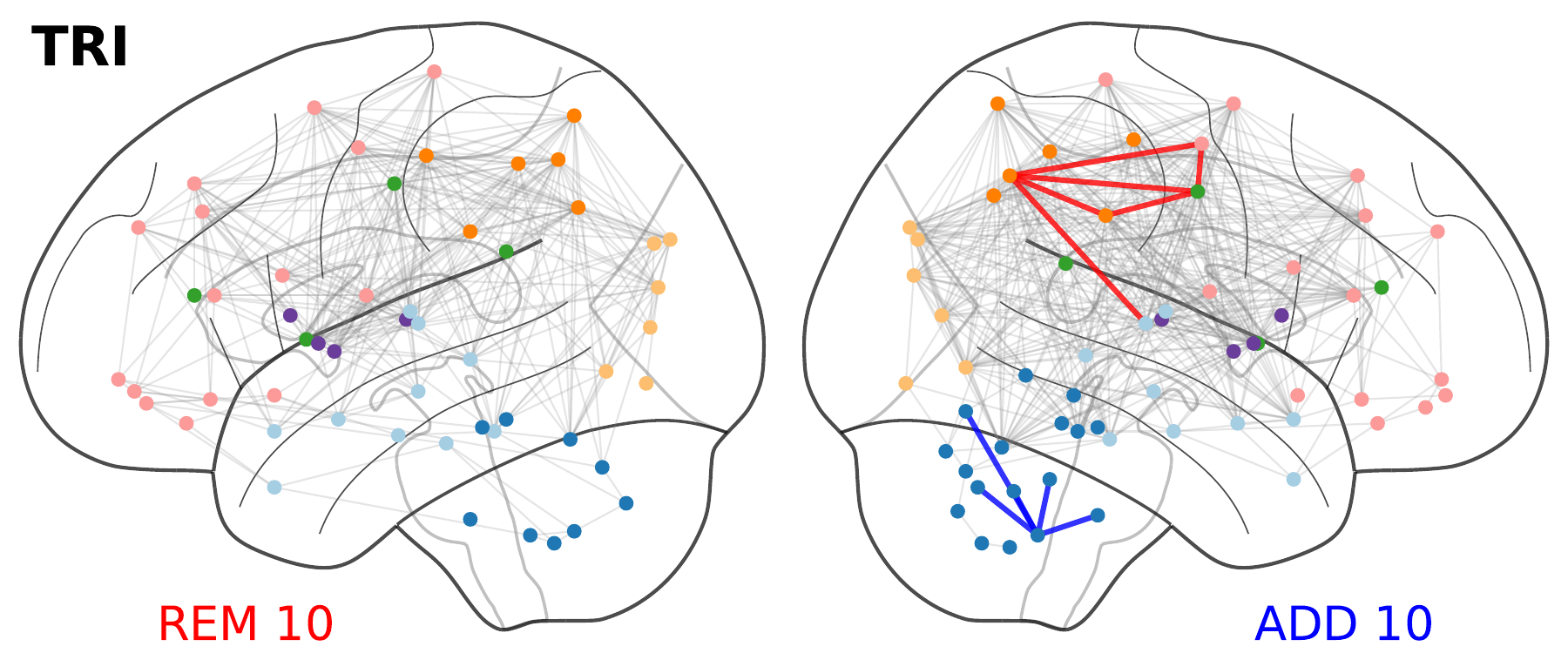}
   \end{minipage}
   \begin{minipage}[b]{0.49\textwidth}
    \centering
    \includegraphics[width=\textwidth]{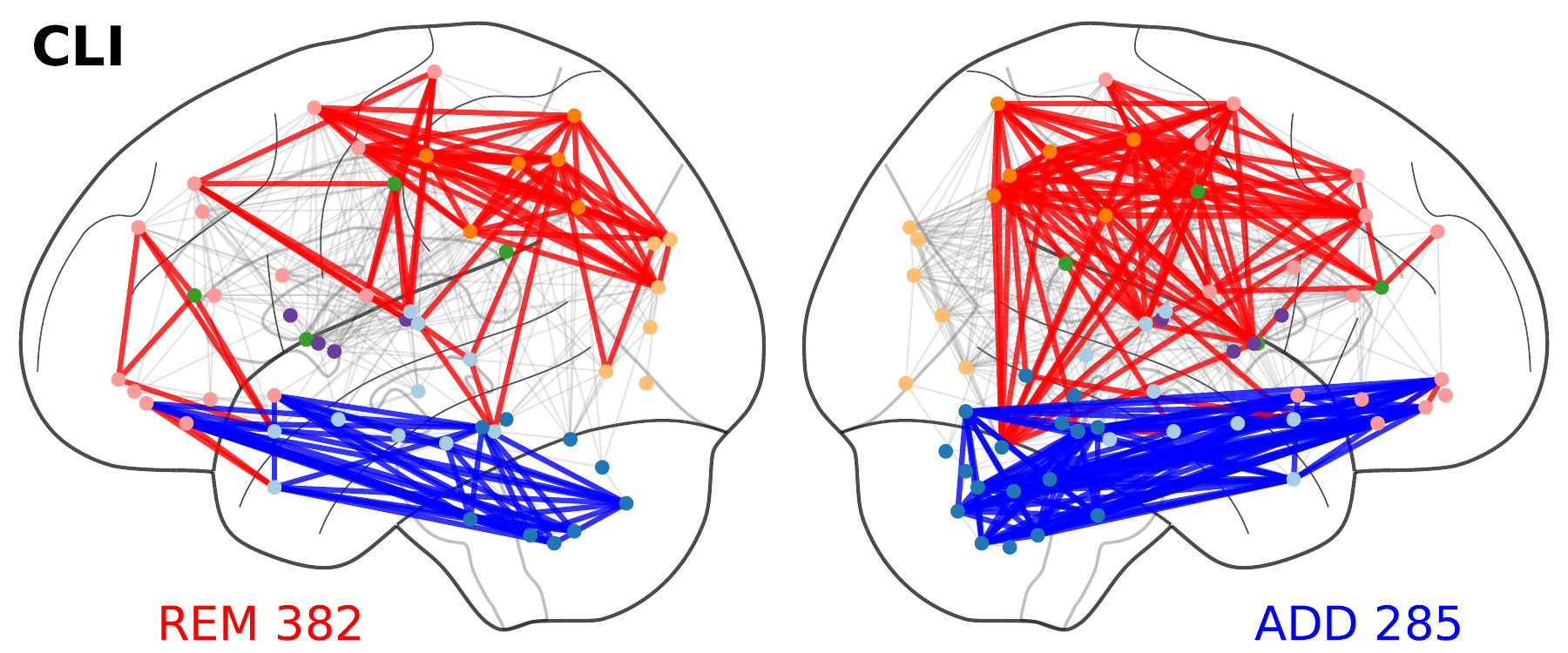}
   \end{minipage}
   \hfill
   \begin{minipage}[b]{0.49\textwidth}
    \centering
    \includegraphics[width=\textwidth]{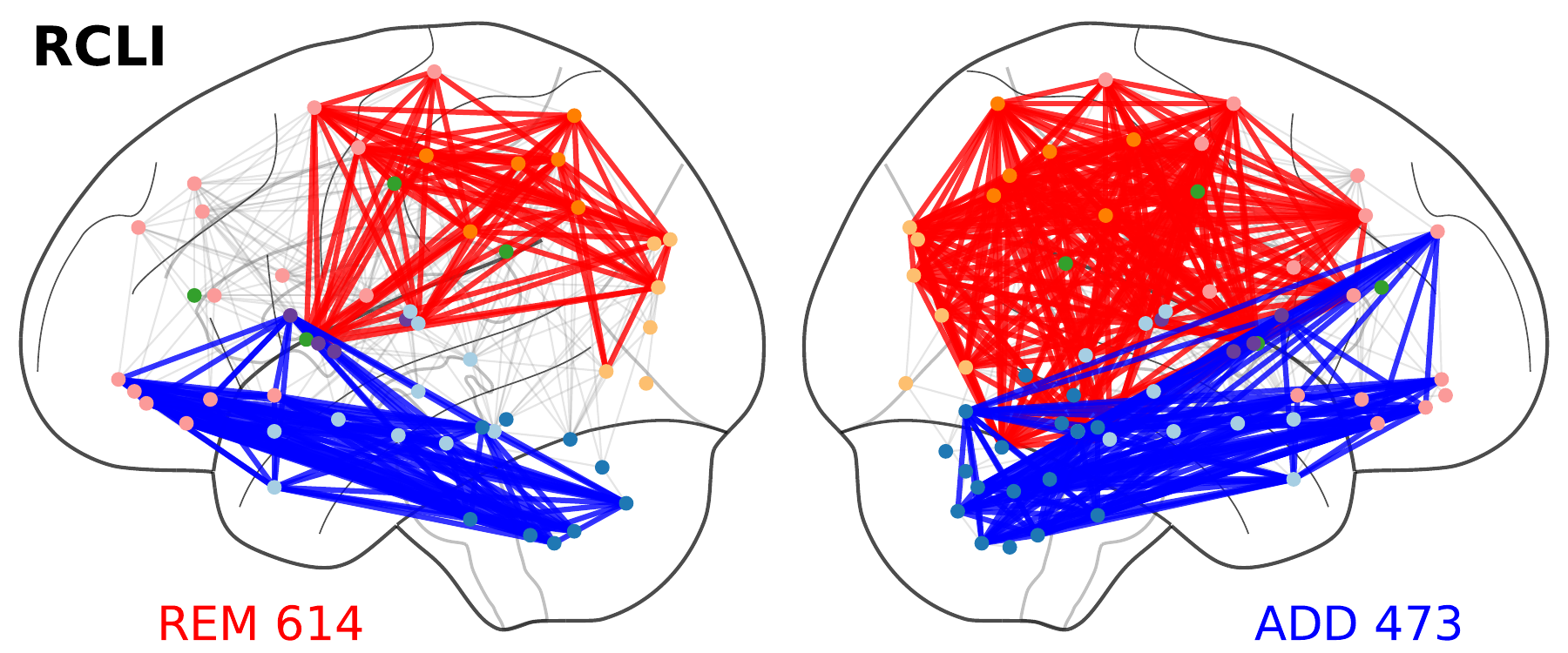}
   \end{minipage}
   \begin{minipage}[b]{0.49\textwidth}
    \centering
    \includegraphics[width=\textwidth]{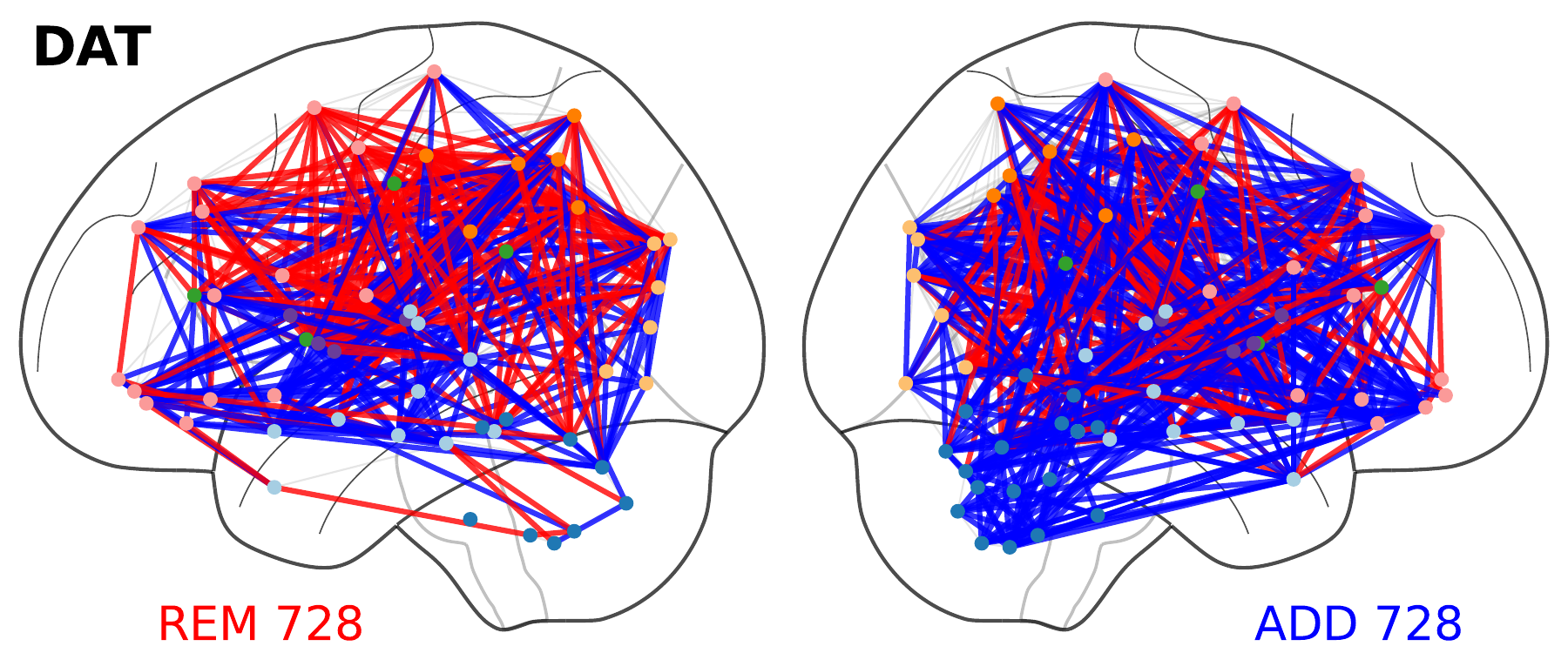}
   \end{minipage}
   \hfill
   \begin{minipage}[b]{0.49\textwidth}
    \centering
    \includegraphics[width=\textwidth]{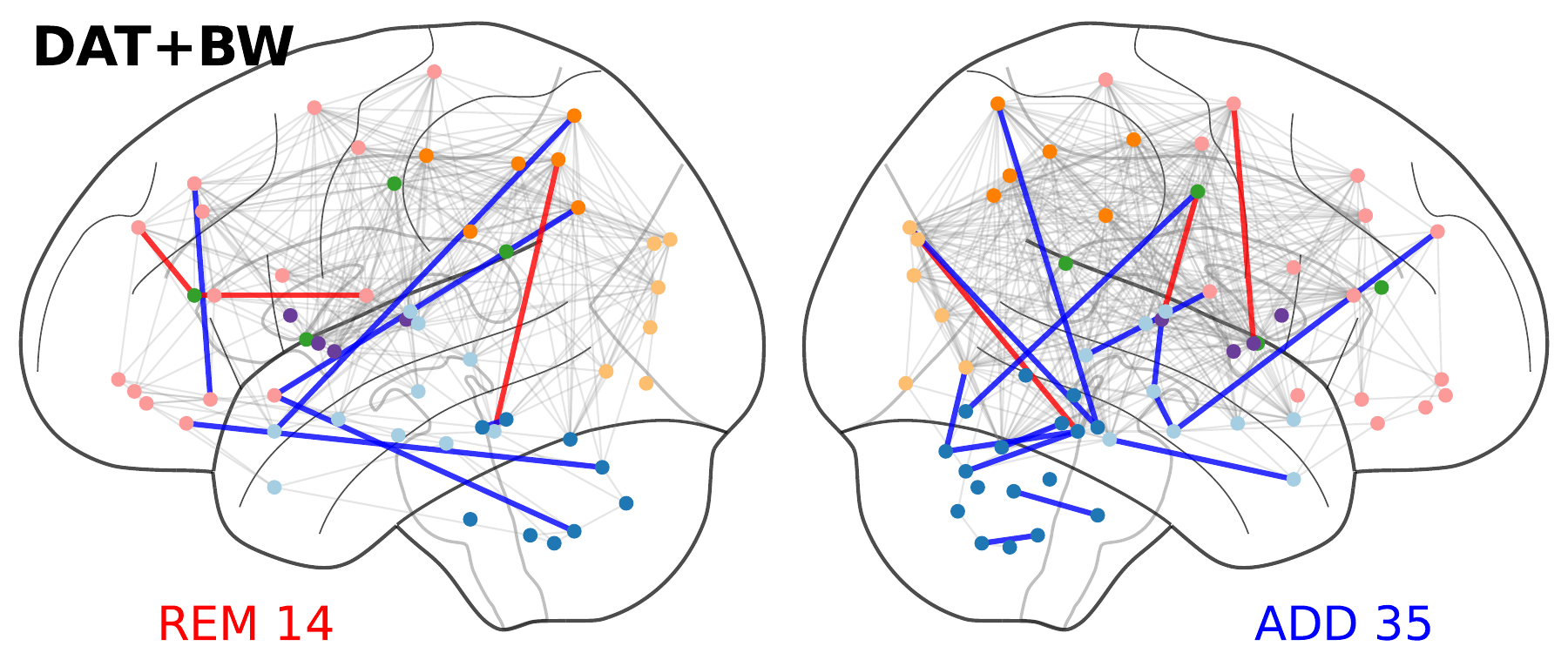}
   \end{minipage}
   \begin{minipage}[b]{0.32\textwidth}
    \centering
    \includegraphics[width=0.8\textwidth]{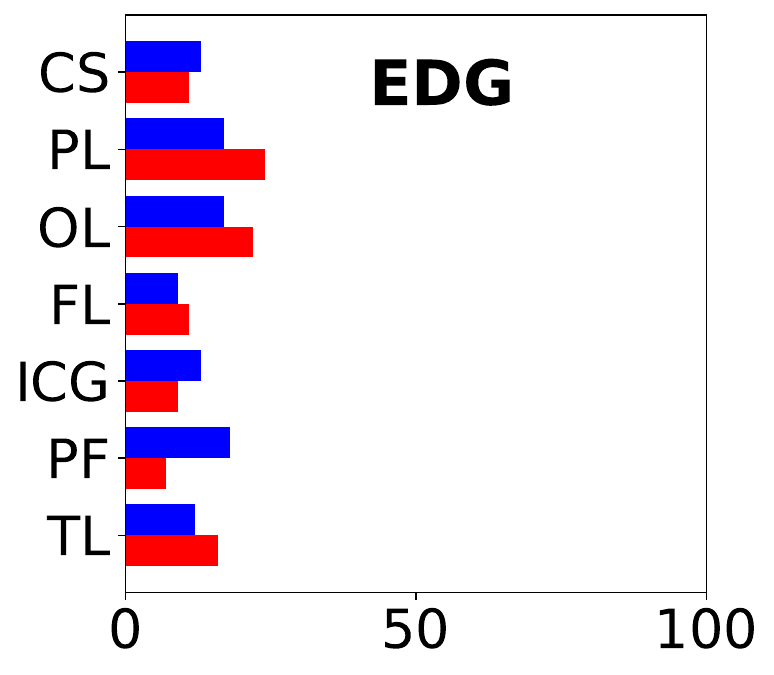}
   \end{minipage}
   \hfill
   \begin{minipage}[b]{0.32\textwidth}
    \centering
    \includegraphics[width=0.8\textwidth]{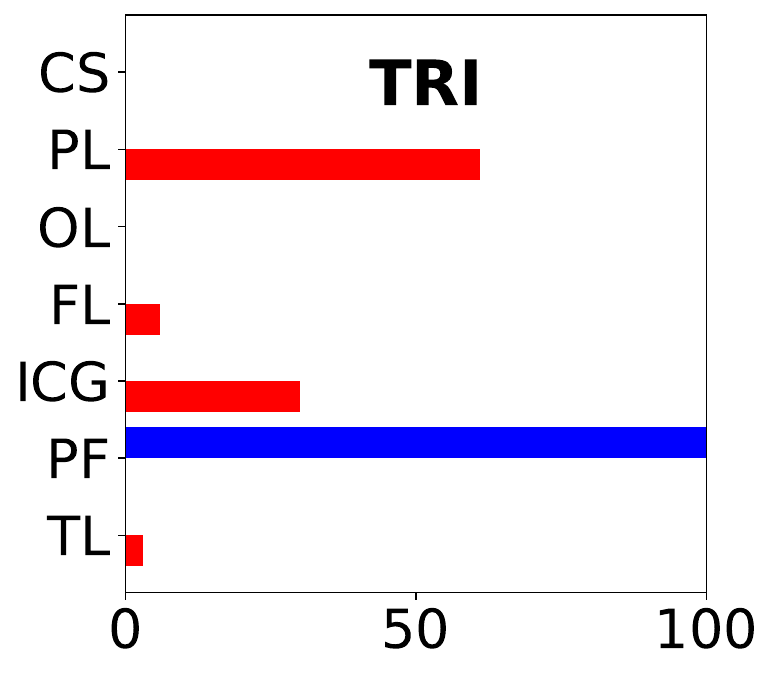}
   \end{minipage}
   \hfill
   \begin{minipage}[b]{0.32\textwidth}
    \centering
    \includegraphics[width=0.8\textwidth]{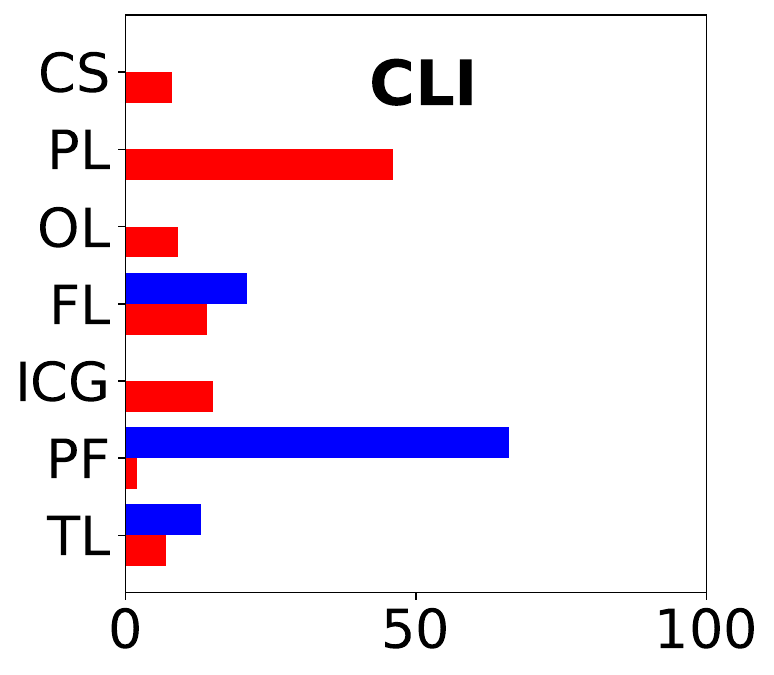}
   \end{minipage}
   \begin{minipage}[b]{0.32\textwidth}
    \centering
    \includegraphics[width=0.8\textwidth]{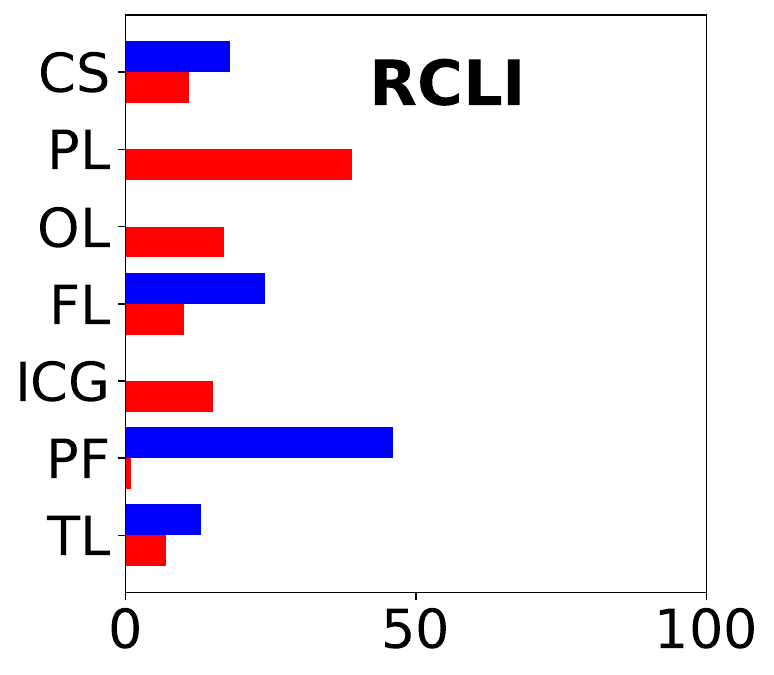}
   \end{minipage}
   \hfill
   \begin{minipage}[b]{0.32\textwidth}
    \centering
    \includegraphics[width=0.8\textwidth]{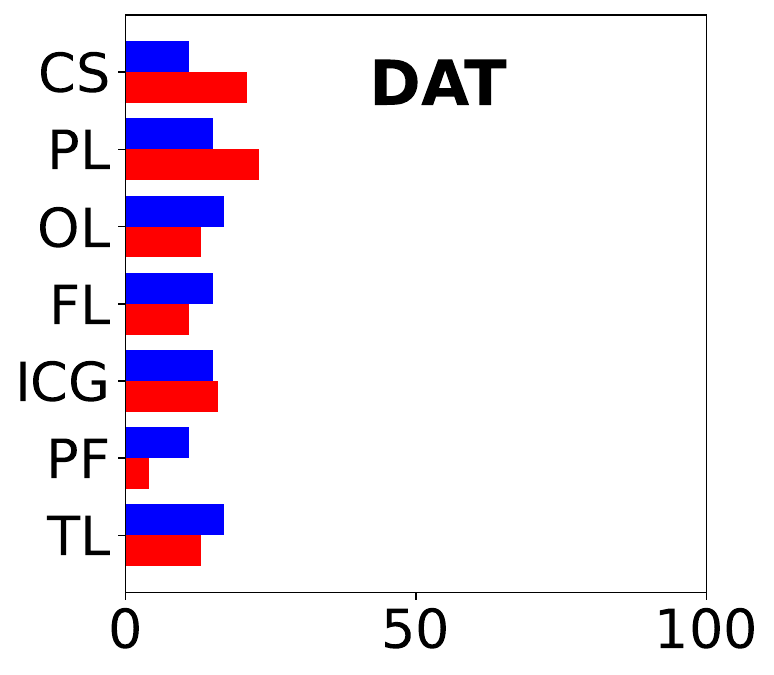}
   \end{minipage}
   \hfill
   \begin{minipage}[b]{0.32\textwidth}
    \centering
    \includegraphics[width=0.8\textwidth]{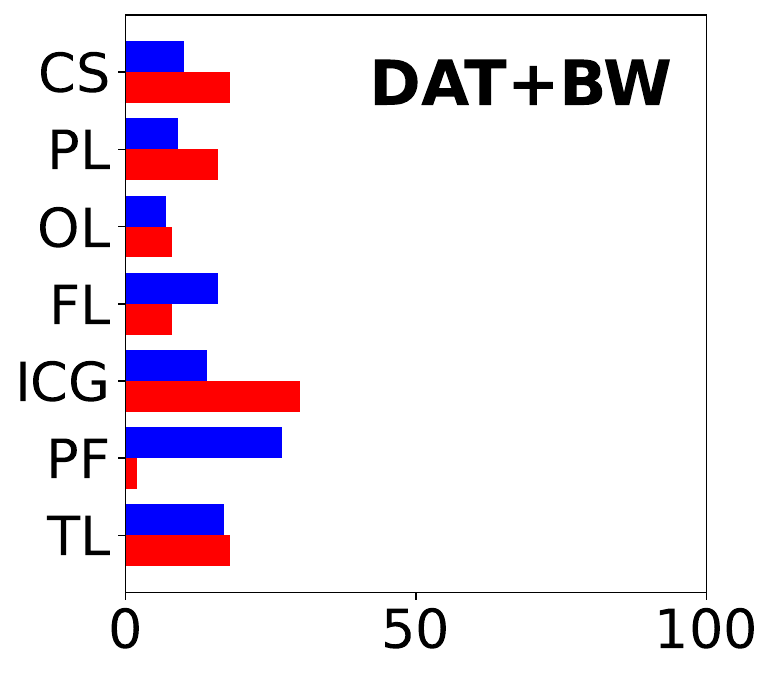}
   \end{minipage}
   \caption{Counterfactual graphs found by each method for patient $9$ in BIP. The upper illustrations show the connectome of the patient where red edges indicate \textit{\color{red}connections to remove}, and blue edges indicate \textit{\color{blue}connections to add} to generate the counterfactual (REM and ADD are the total numbers). The lower barplots show to which brain lobes the ROIs of the edges added (in blue) and removed (in red) belong (as percentages).}
   \label{fig:example}
   \vspace{8mm}
\end{figure}

\mpara{BIP Dataset.}
\Cref{fig:example} shows the counterfactual graphs and the distributions of edges changed among the brain lobes, for patient $9$ in the BIP dataset.
This patient is classified as ``Typically Developed''.
This example serves to confirm the effectiveness of \tri and \cli in generating more compact and interpretable explanations. Specifically, \tri produces a counterfactual graph that closely resembles the input network, with only 10 edges added and 10 edges removed. On the other hand, \cli concentrates its changes in two specific regions: the Parietal Lobe (with connections removed) and the Posterior Fossa (with connections added).
The output of \tri can be summarized by the following simple \textit{counterfactual statement}:
\begin{displayquote}
\emph{Patient X is classified as Typically Developed. If X's brain had less activation in the \textsc{\color{red} Parietal Lobe} and the \textsc{\color{red} Insula Cingulate Gyri}, and more co-activation in the \textsc{\color{blue} Posterior Fossa}, then X would have been classified as Bipolar.}
\end{displayquote}
Given the Cerebellum's crucial role in emotional regulation, it's worth noting that the Posterior Fossa, which houses the Cerebellum, is an important area of study in bipolar disorder research~\cite{ewald2022posterior,kim2013posterior,minichino2014role}.

In contrast, \edg and \dataset generate sparser explanations that involve all the brain regions, making them more complex. The same is true for the backward search method (\datasetbw).
Based on these results, we can conclude that the baseline methods are less effective at producing counterfactual explanations that are consistent with the terminology used to describe the organization of the brain.

\begin{figure}[t!]
\centering
    \begin{minipage}[b]{\textwidth}
    \centering
    \includegraphics[clip,width=.9\textwidth]{Figures/brains/legend.pdf}
    \end{minipage}
    \begin{minipage}[b]{0.63\textwidth}
    \centering
    \includegraphics[width=\textwidth]{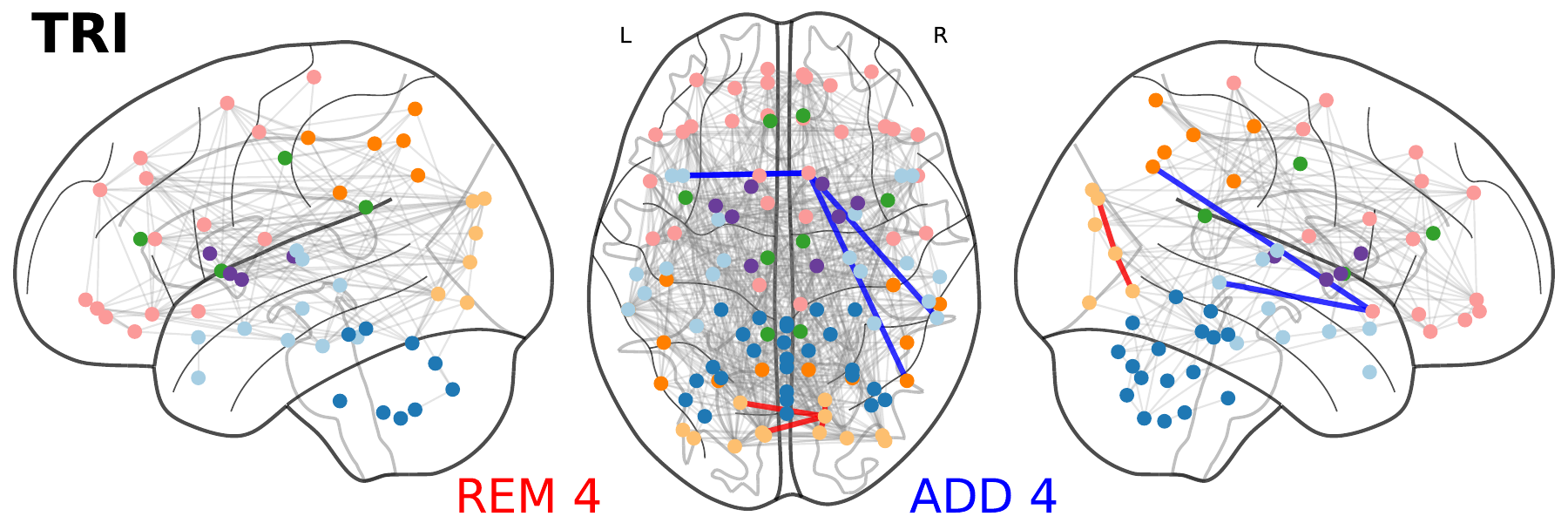}
   \end{minipage}
   \hfill
   \begin{minipage}[b]{0.30\textwidth}
    \centering
    \includegraphics[width=\textwidth]{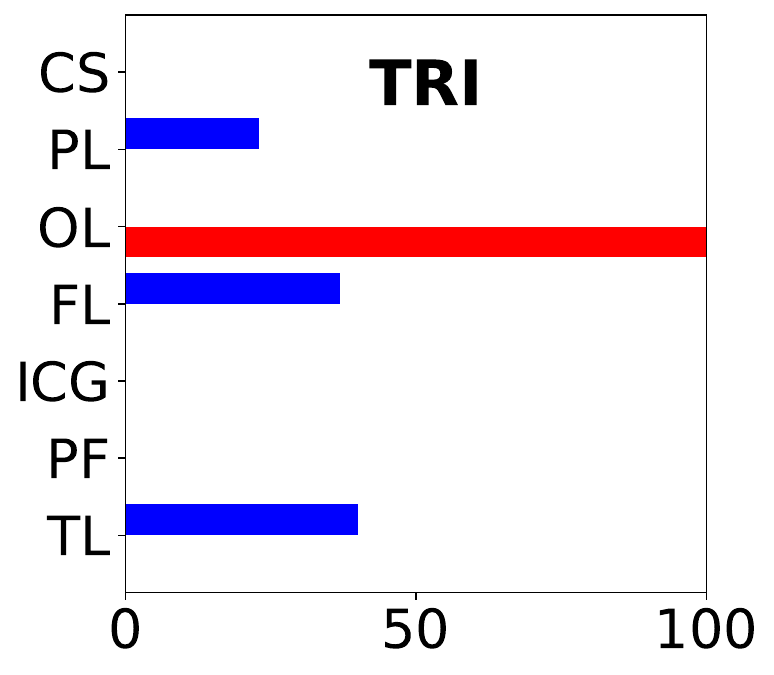}
   \end{minipage}
   \begin{minipage}[b]{0.63\textwidth}
    \centering
    \includegraphics[width=\textwidth]{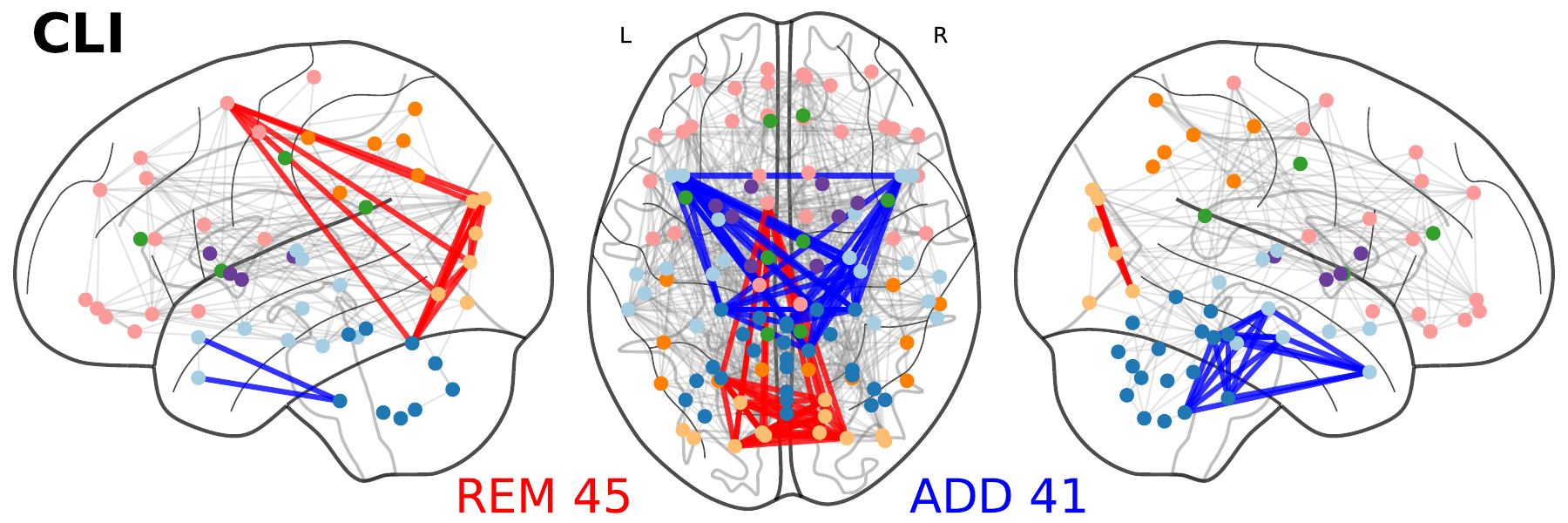}
   \end{minipage}
   \hfill
   \begin{minipage}[b]{0.30\textwidth}
    \centering
    \includegraphics[width=\textwidth]{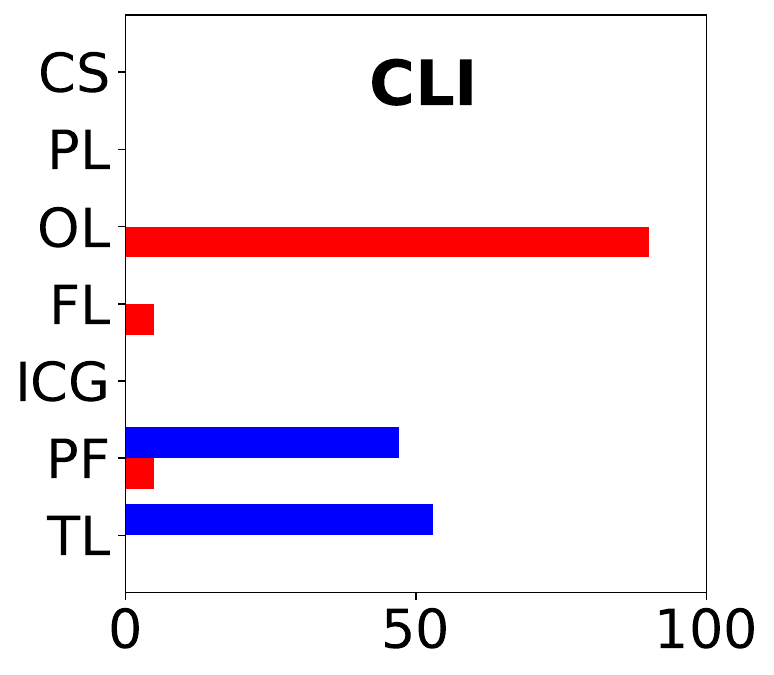}
   \end{minipage}
   \caption{Counterfactual graphs found by each method for patient $22$ in ADHD. The left illustrations show the connectome of the patient where red edges indicate \textit{\color{red}connections to remove}, and blue edges indicate \textit{\color{blue}connections to add} to generate the counterfactual (REM and ADD are the total numbers). The right barplots show to which brain lobes the ROIs of the edges added (in blue) and removed (in red) belong (as percentages).}
   \label{fig:example2}
\end{figure}

\mpara{ADHD Dataset.}
As a last example, \Cref{fig:example2} depicts the counterfactual graphs and the distributions of edges changed among the brain lobes, for patient $22$ in the ADHD dataset, who is classified by ADHD.
In this case, the counterfactuals generated by \tri and \cli are quite similar, with edges removed from the Occipital Lobe and added primarily in the Temporal Lobe. Additionally, \cli adds connections in the Posterior Fossa, while \tri adds them in the Frontal Lobe. It's worth noting that several studies on the structural and functional neuroimaging of ADHD patients have shown alterations in Occipital Regions~\cite{wu2020role,soros2017inattention}.
The output of \cli can be summarized by the following simple \textit{counterfactual statement}:
\begin{displayquote}
\emph{Patient X is classified as ADHD. If X's brain had less activation in the \textsc{\color{red} Occipital Lobe}, and more co-activation in the \textsc{\color{blue} Posterior Fossa} and the \textsc{\color{blue} Temporal Lobe}, then X would have been classified as Typically Developed.}
\end{displayquote}

\begin{figure}[t!]
   \vspace{-2mm}
   \begin{minipage}[b]{0.49\textwidth}
    	\includegraphics[width=\textwidth]{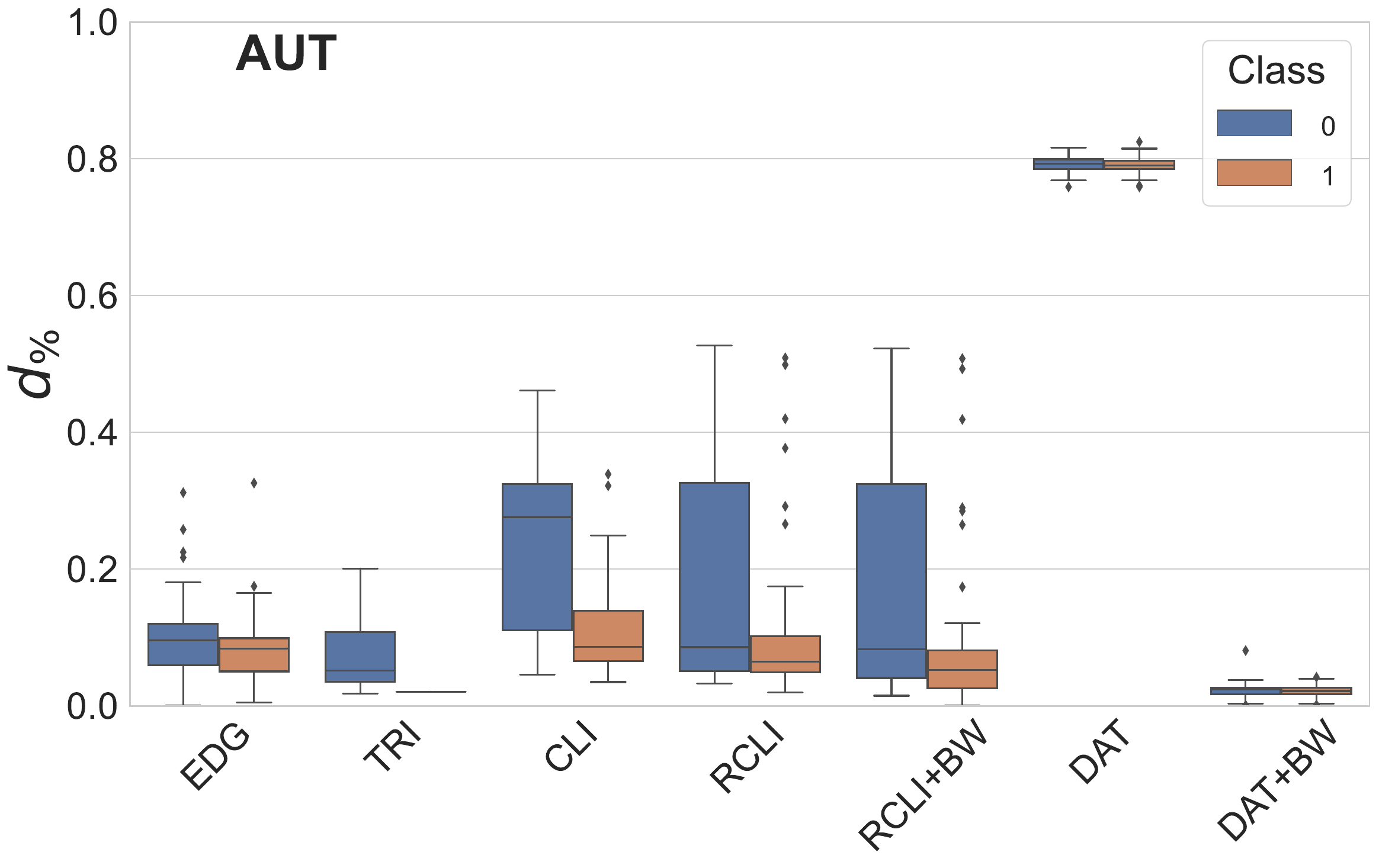}
   \end{minipage}
   \hfill
   \begin{minipage}[b]{0.49\textwidth}
    	\includegraphics[width=\textwidth]{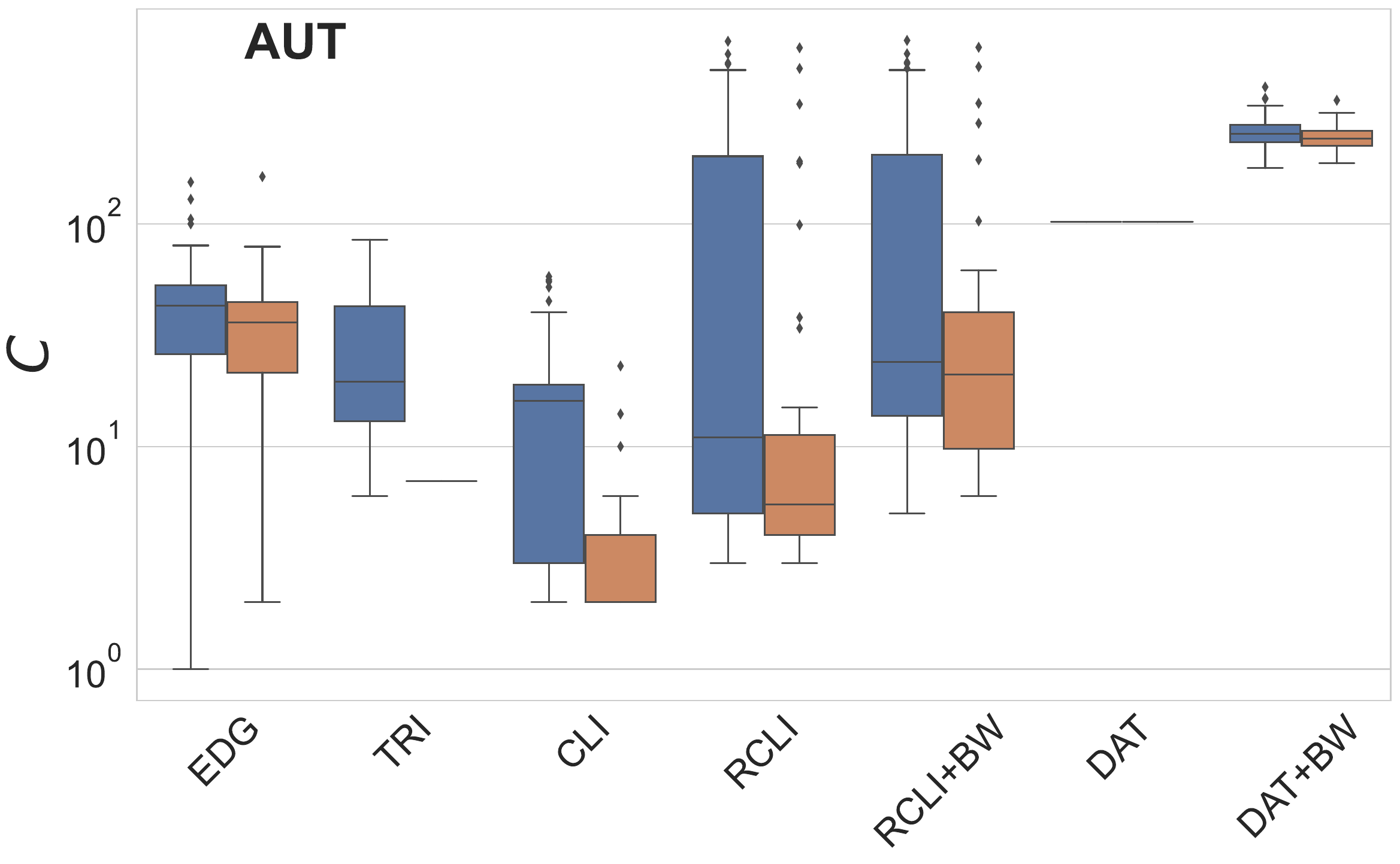}
   \end{minipage}
   \begin{minipage}[b]{0.49\textwidth}
     \includegraphics[width=\textwidth]{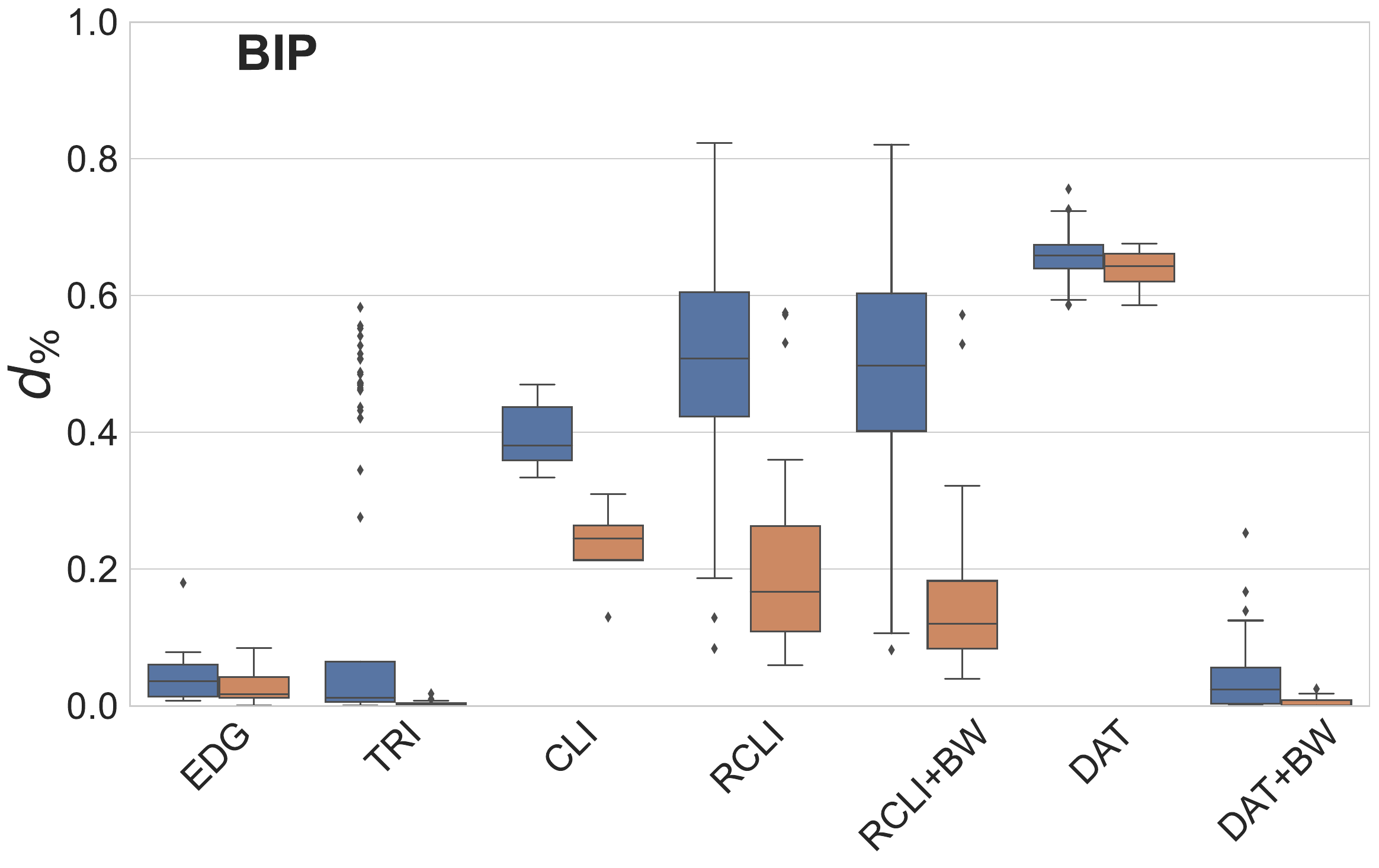}
   \end{minipage}
   \hfill
   \begin{minipage}[b]{0.49\textwidth}
    	\includegraphics[width=\textwidth]{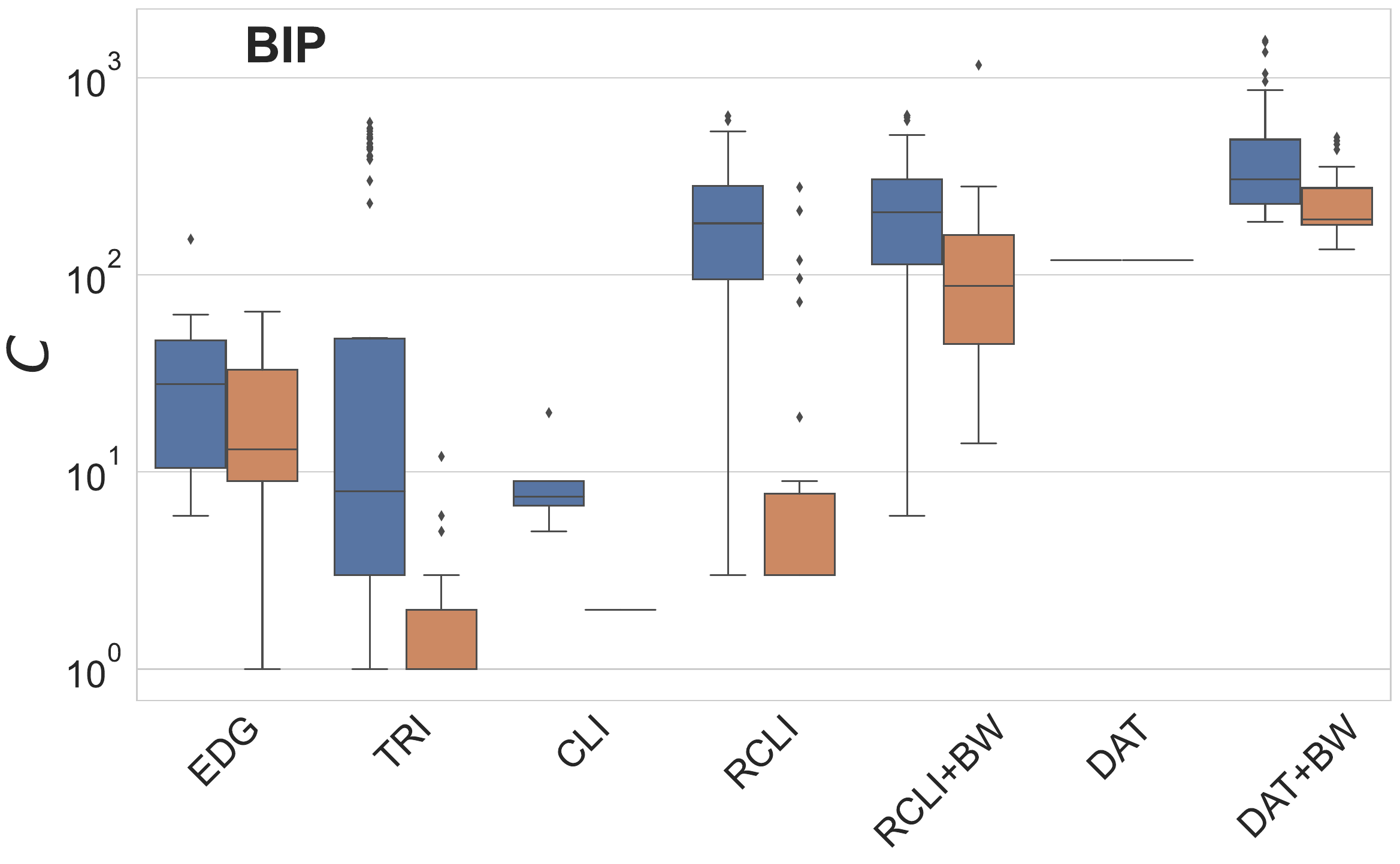}
   \end{minipage}
   \begin{minipage}[b]{0.49\textwidth}
    	\includegraphics[width=\textwidth]{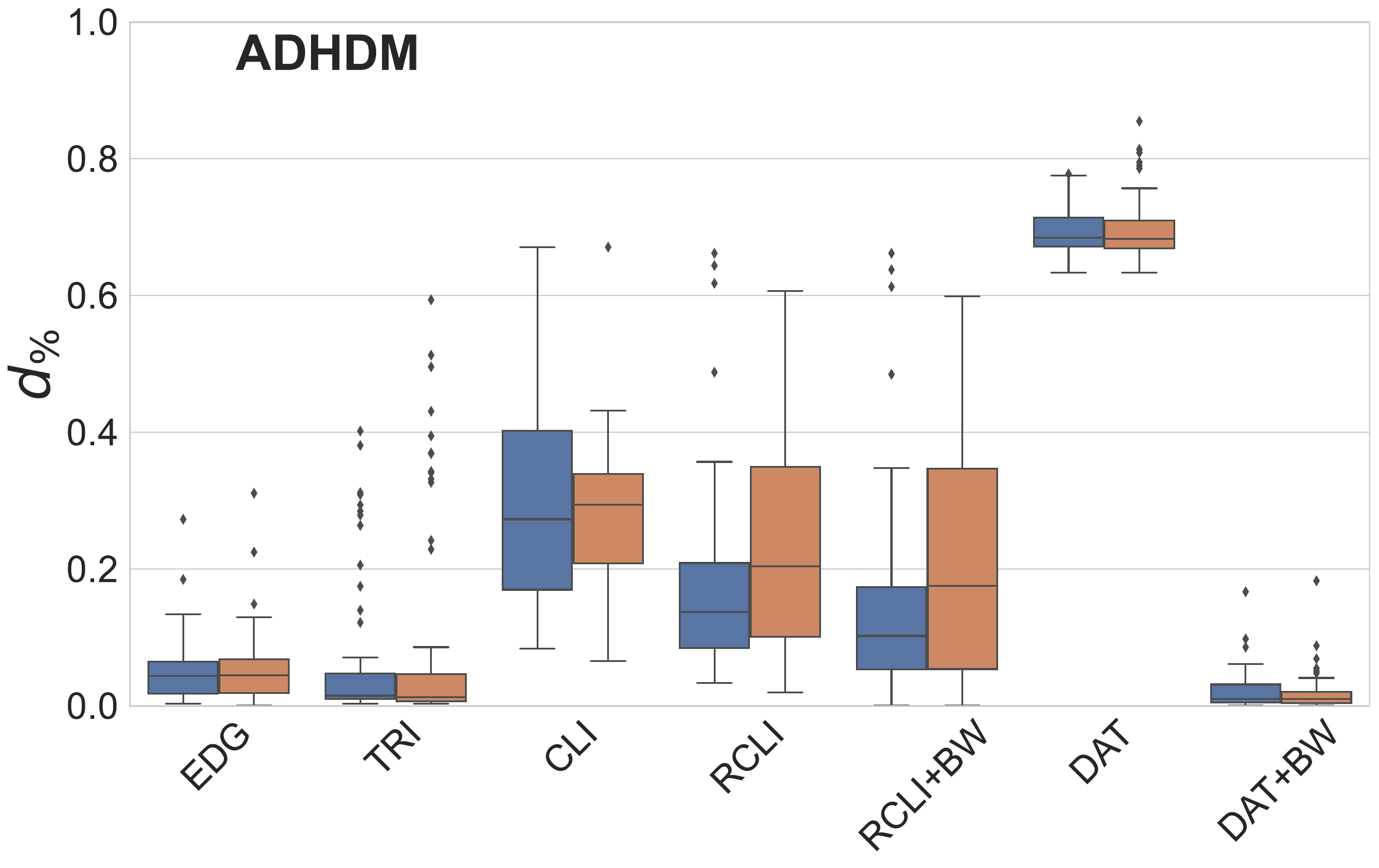}
   \end{minipage}
   \hfill
   \begin{minipage}[b]{0.49\textwidth}
    	\includegraphics[width=\textwidth]{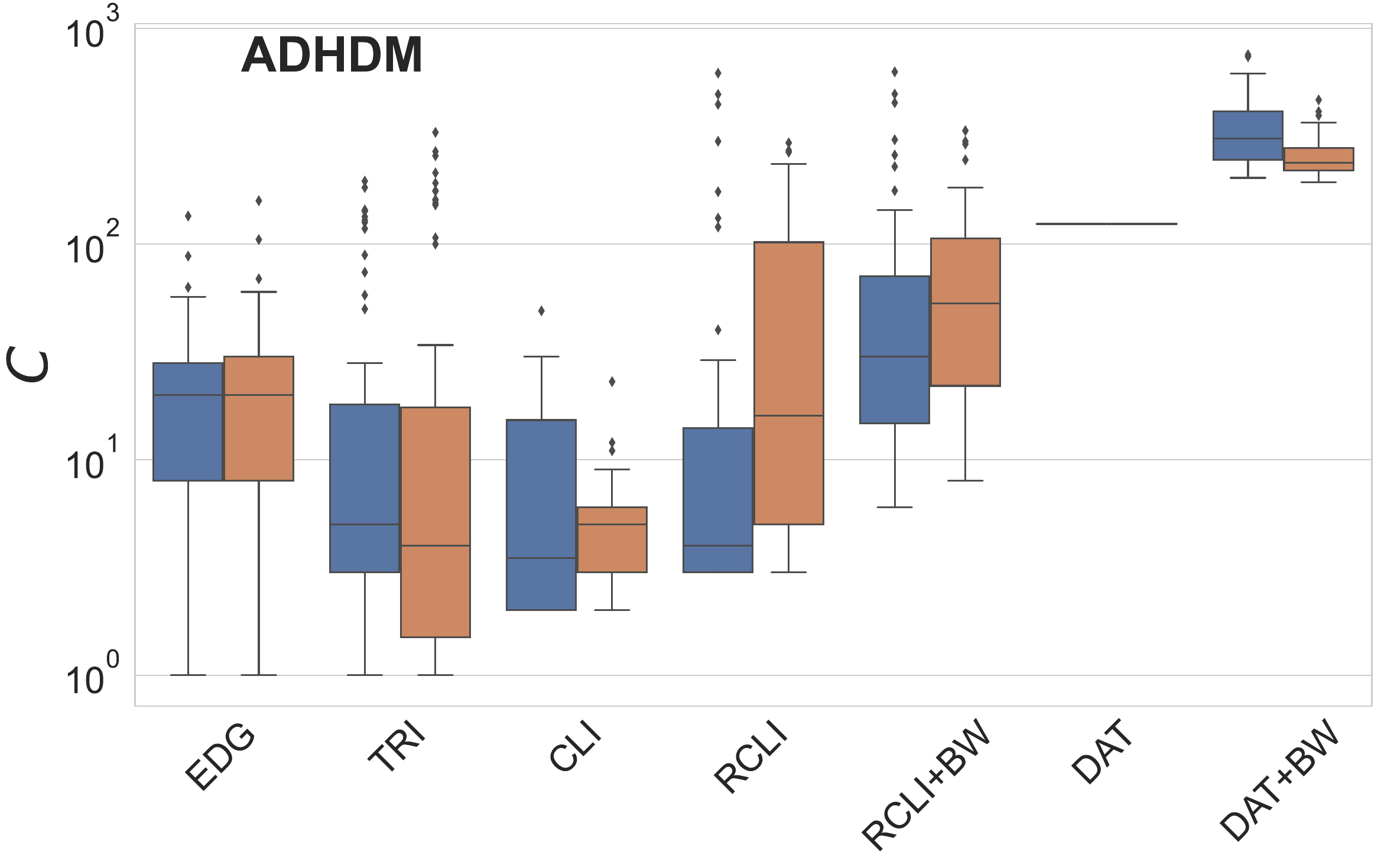}
   \end{minipage}
\vspace{-4mm}
   \caption{Distribution of symmetric differences between original and counterfactual graphs ($d_{\%}$) and of calls to the oracle ($C$), for each counterfactual generation method, for AUT (1st row), BIP (2nd row), and ADHDM (3rd row).}
   \label{fig:distance_methods}
   \vspace{2mm}
\end{figure}

\subsection{Quantitative Comparison}\label{sec:global}
We next present a comparison of the various counterfactual generation methods, using the metrics outlined in \Cref{sec:metrics}.

\Cref{fig:distance_methods} reports the distribution of $d_{\%}$ and $C$ values (the latter in logarithmic scale) at the class level for each method across three datasets (AUT, BIP, and ADHDM). Results for the remaining datasets can be found in the supplementary material shared in our repository.
Computation complexity, which is measured as the number of calls to the black-box classifier, varies across the different methods tested, with \dataset and \datasetbw being the most time-consuming due to the need to compare the input network with each graph classified in the opposite class. We note that \datasetbw requires slightly more calls to the oracle $C$ because it also performs a backward search.
\cli tends to find solutions more quickly than the other methods due to its tendency to make larger changes in the regions of the network, whereas \edg and \tri may require more iterations (and thus calls to the oracle) to achieve the same change.

We next examine the proximity of the counterfactual graphs to the corresponding input networks. As we observed in the previous section, the explanations generated by \dataset differ significantly from the input networks, as it searches for counterfactuals among the graphs in the dataset (which can vary considerably from each other) rather than perturbing the network itself.
The application of the backward search on top of \dataset results in counterfactuals that are much closer to the original networks compared to other methods. 
Interestingly, applying a backward search after \rcli does not significantly alter the resulting counterfactuals, suggesting that these solutions are more robust than those generated by \dataset.
Finally, since both \tri and \edg change a few edges at each iteration, the corresponding distributions of symmetric differences are comparable.

Methods such as \cli and \rcli operate on the maximal cliques in the network, which causes them to change a larger number of edges at each iteration, resulting in counterfactuals that are more distant than those obtained by \edg and \tri.
It is important to stress that \cli and \rcli, by design, are expected to induce larger changes when producing a counterfactual as they use a coarser-grain vocabulary in the explanation (dense regions), w.r.t. the fine-grain approaches of \edg and \tri. As motivated in \Cref{sec:intro}, we aim to have explanations at the level of regions (in which the changes are concentrated), because these are more interpretable for the domain expert than a simple list of flipped edges.


We finally report the flip rate per class (class 0/class 1) for each dataset (columns) and each method (rows) in \Cref{tab:fr}. 
We remind that \rcli was tested only in the datasets where the brains' parcellations were available (i.e., all but OHSU, PEK, and KKI).
By definition, the flip rate of \dataset is always 100\%, as it picks the closest counterfactual among the graphs in the database.
The other methods, instead, did not achieve a perfect score, as they were run for a fixed number of iterations.
In particular, \tri is run for at most $\min\left(|E_-|, |E_+|\right)$ iterations, \cli and \rcli for at most $\max_I =200$ iterations, and \edg for st most 2000 iterations. 
We observe that \tri converges to a counterfactual more frequently than the other methods, even in the unbalanced ADHD dataset. However, it struggles in the three sparsest networks (OHSU, PEK, KKI), likely because triadic closure is less observable in these graphs, while \edg adds and removes edges more indiscriminately, which allows it to eventually find a counterfactual even in these cases. 
Finally, \cli strikes a balance between \edg and \tri, as it acts on maximal cliques and can thus remove and add cliques even in sparser graphs (where cliques are just edges).

\begin{table}[t!]
    \centering
    \caption{Flip rate (class 0/ class 1), for each method and each dataset. Flip rate for \datasetbw and \rclibw are not reported as they are the same as \dataset and \rcli.}
    \vspace{-2mm}
    \label{tab:fr}
    \begin{tabular}{cccccccc}
    \toprule
    \textbf{Method} &
    \textbf{AUT} & 		
    \textbf{BIP} & 		
    \textbf{ADHD} & 		
    \textbf{ADHDM} & 		
    \textbf{OHSU} & 		
    \textbf{PEK} &
    \textbf{KKI} \\
    \midrule
    EDG & 100/85  & 70/100  & 75/100  & 74/100  & 100/87  & 100/91  & 90/100 \\
    TRI & 100/100  & 100/100  & 100/100  & 98/100  & 62/58  & 89/96  & 90/71 \\
    CLI & 91/100  & 100/100  & 100/100  & 56/100  & 53/87  & 100/91  & 90/100 \\
    RCLI & 96/93  & 94/100  & 95/98  & 61/100  & -  & - & -\\
    DATA & 100/100  & 100/100  & 100/100  & 100/100  & 100/100  & 100/100  & 100/100 \\
    \bottomrule
    \end{tabular}
    \vspace{2mm}
\end{table}
\section{Conclusions and Future Work} \label{sec:conclusions}
We introduced a general framework, called \emph{density-based counterfactual search} (\dcs), for generating instance-level counterfactual explanations for graph classifiers using the alteration of dense substructures. 
This framework identifies the most informative regions of the graphs and manipulates them by adding or removing dense structures until a counterfactual is found. 
The modularity of the framework allows users to customize their counterfactual search based on their specific needs.
We instantiated \dcs in two special cases: \tri and \cli. In \tri, we search for counterfactual graphs by opening or closing triangles, while in \cli, we move to a counterfactual search driven by maximal cliques.
Additionally, we showed a variation of \cli, called \rcli, which leverages the brain's parcellation to rank the nodes and encourage changes within the same lobes of the brain. This variation generates more interpretable explanations for brain networks.

As further work, we plan to address the feasibility and robustness constraints, pivotal in many counterfactual search scenarios. The feasibility constraint arises because, for certain types of data, some counterfactuals may not be feasible or may not exist at all. For example, not all the counterfactuals generated for molecule graphs may be chemically feasible structures. On the other hand, robustness to noise, i.e., when small perturbations to the counterfactual do not change its predicted class, makes the counterfactual explanation more trustworthy and is thus a desirable characteristic.

\clearpage
\bibliographystyle{splncs04}
\bibliography{bibfile,biblio}

\end{document}


\title{Supplementary Material for Counterfactual Explanations for Graph Classification Through the Lenses of Density}

\maketitle \sloppy              

\appendix 
\section{Other Results}
\begin{figure}[h]
   \begin{minipage}[b]{0.49\textwidth}
        \centering
        \includegraphics[width=.86\textwidth]{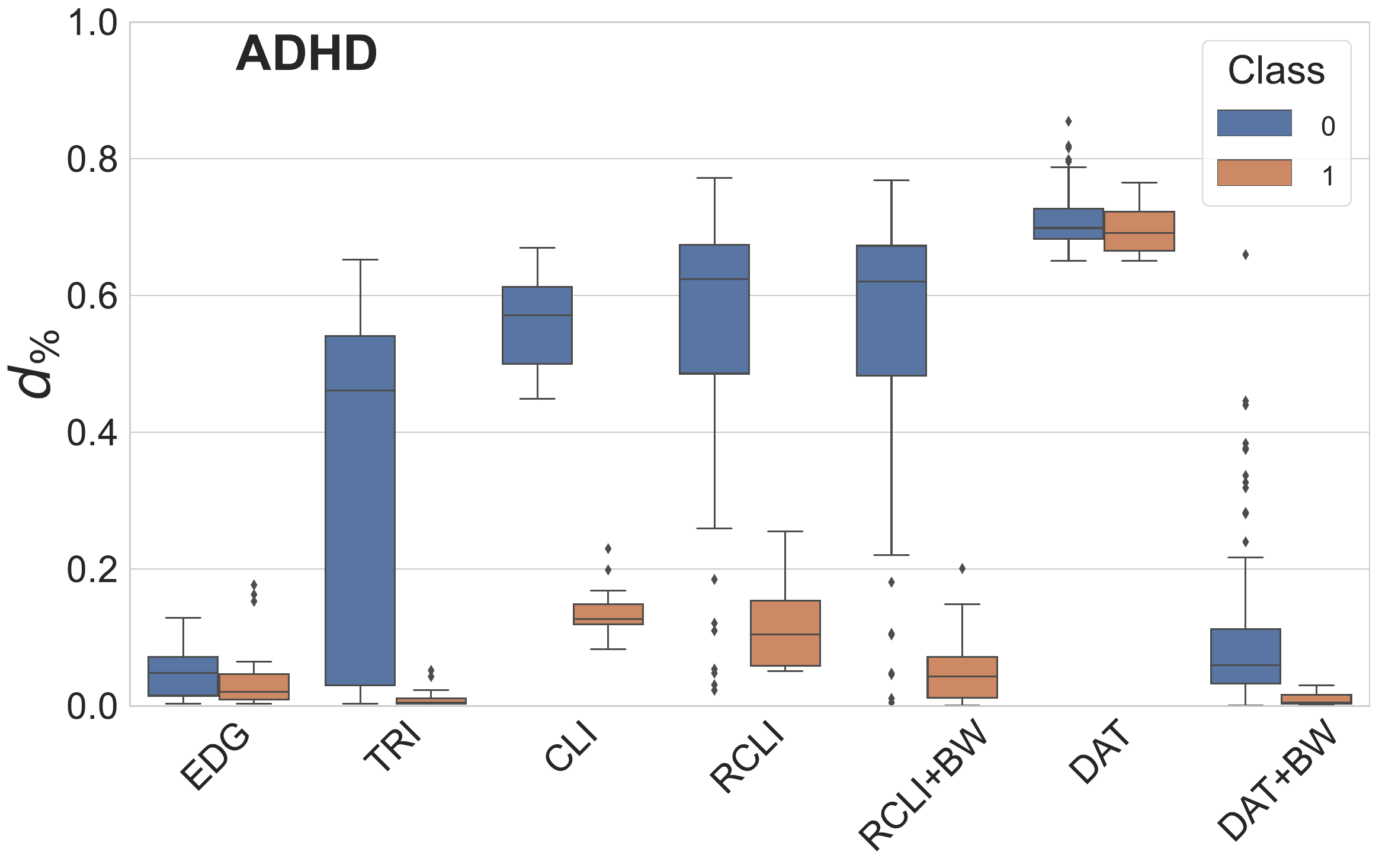}
   \end{minipage}
   \begin{minipage}[b]{0.49\textwidth}
        \includegraphics[width=.86\textwidth]{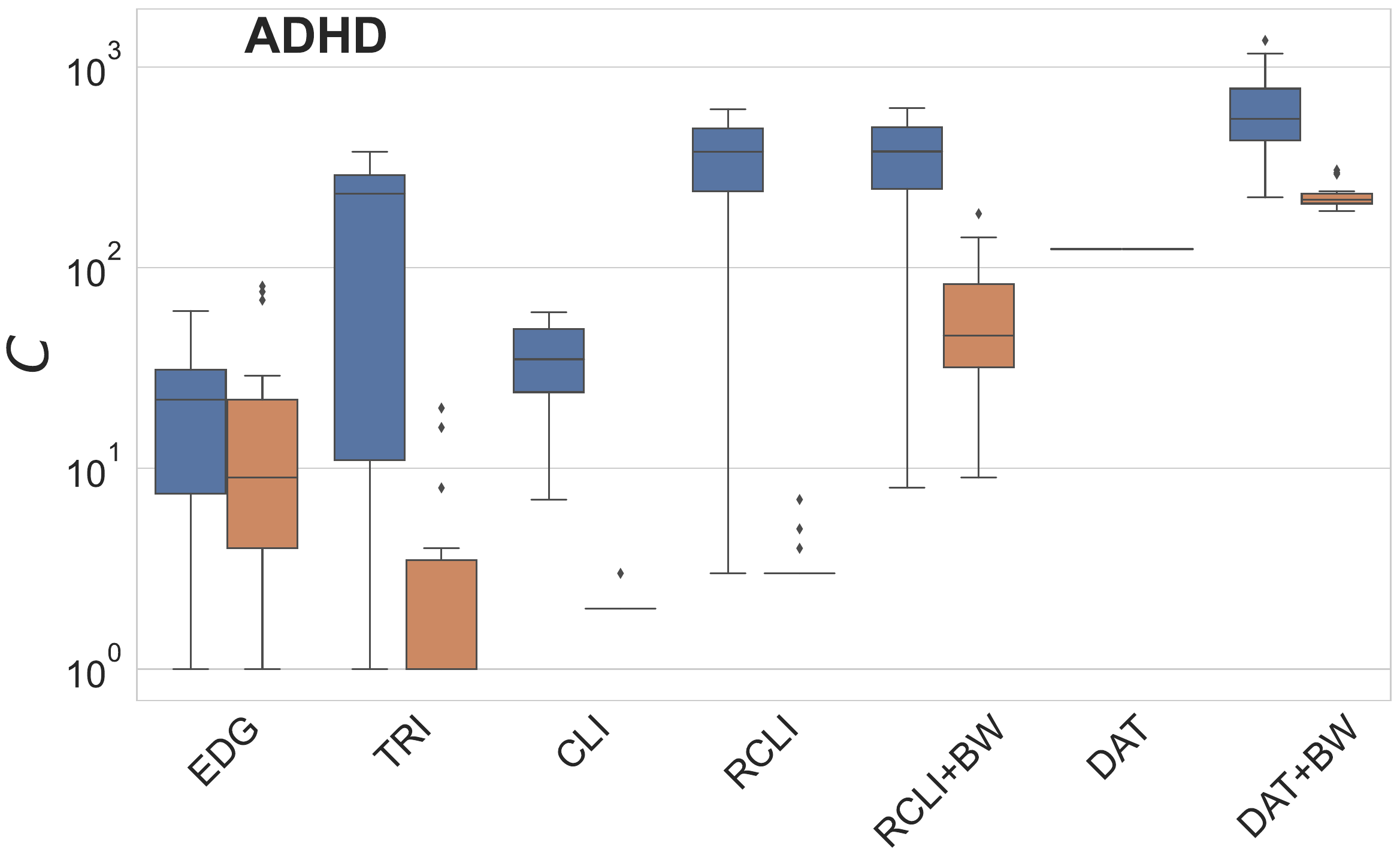}
   \end{minipage}
   \begin{minipage}[b]{0.49\textwidth}
        \includegraphics[width=.86\textwidth]{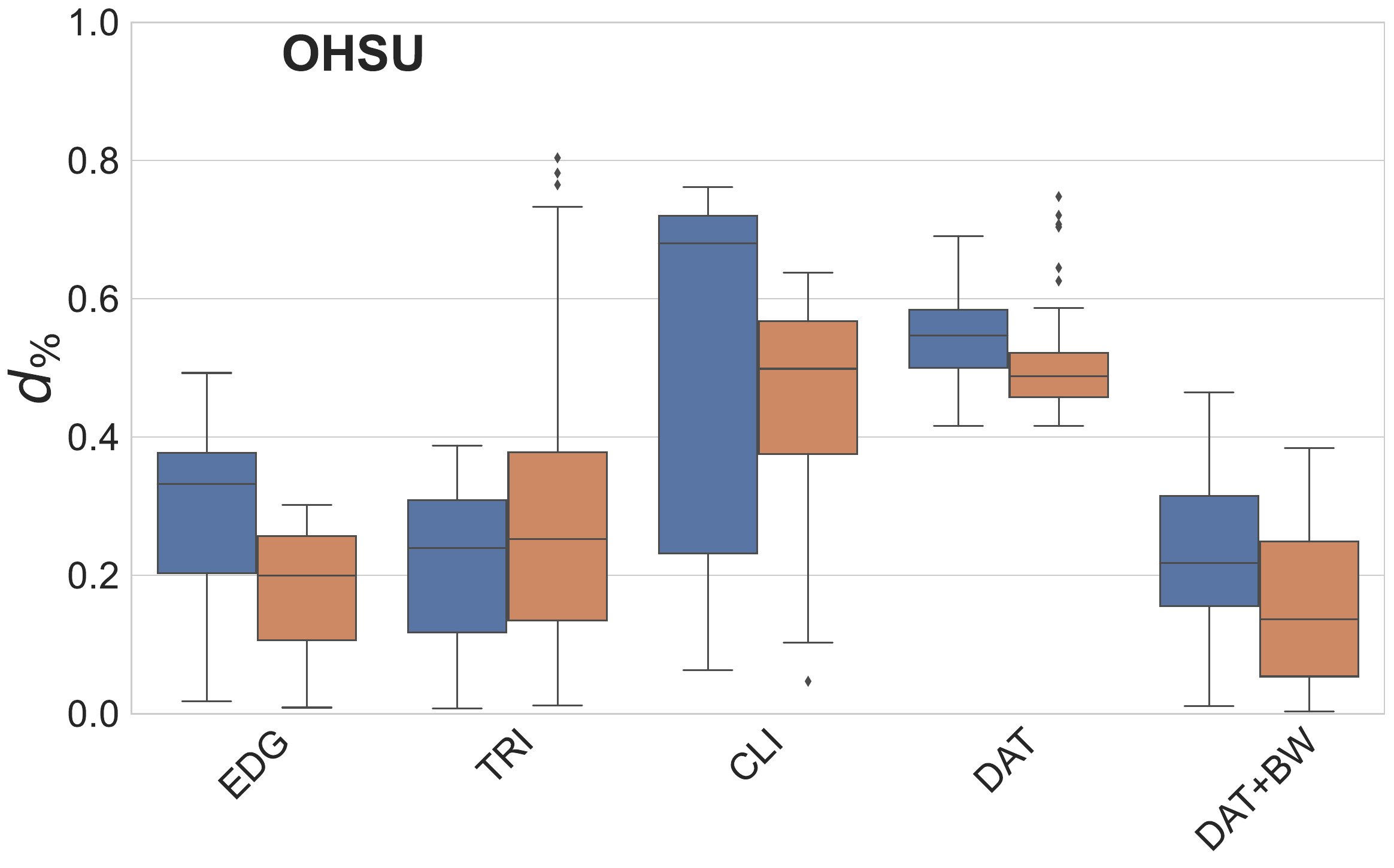}
   \end{minipage}
   \begin{minipage}[b]{0.49\textwidth}
        \includegraphics[width=.86\textwidth]{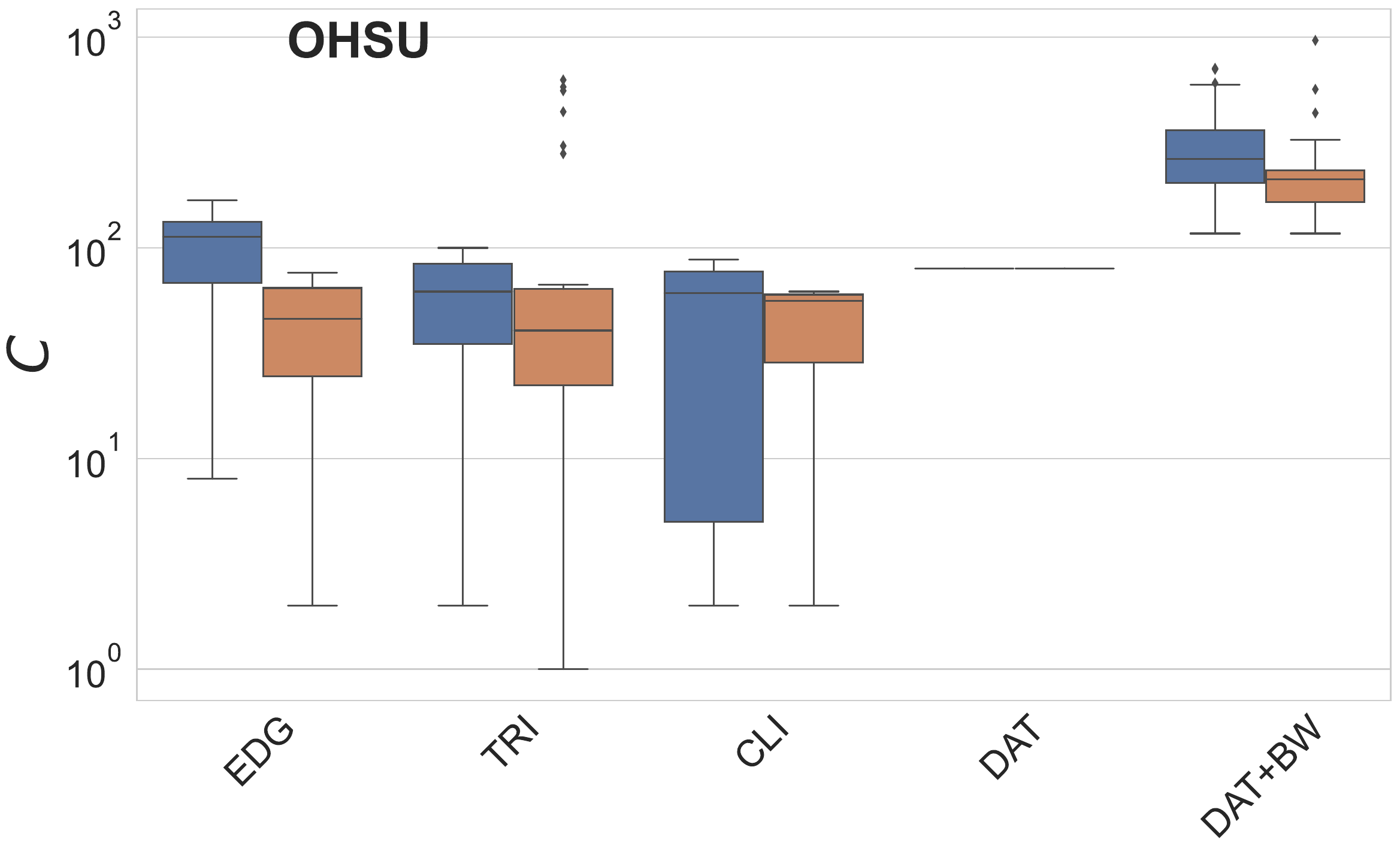}
   \end{minipage}
   \begin{minipage}[b]{0.49\textwidth}
	\includegraphics[width=.86\textwidth]{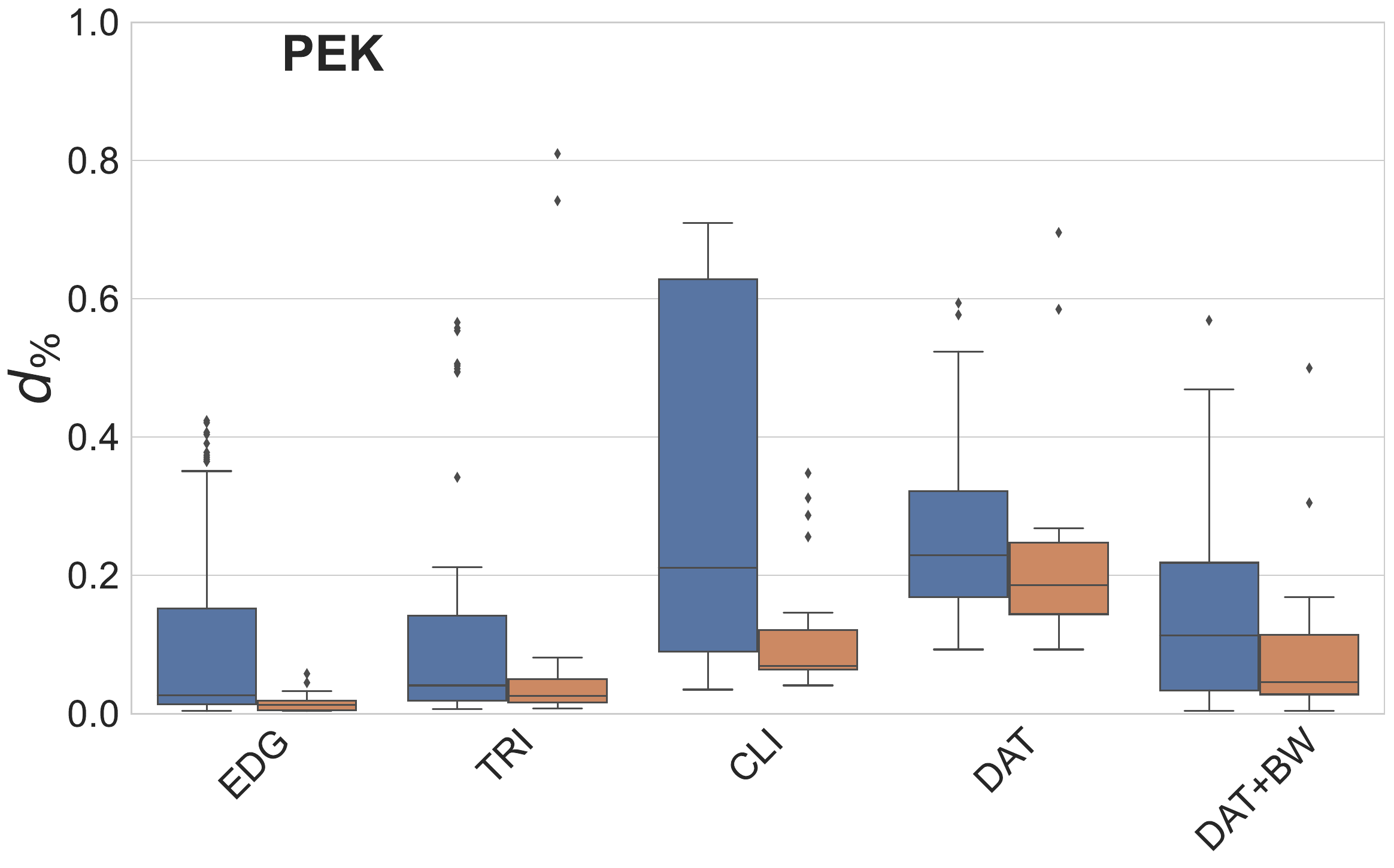}
   \end{minipage}
   \begin{minipage}[b]{0.49\textwidth}
        \includegraphics[width=.86\textwidth]{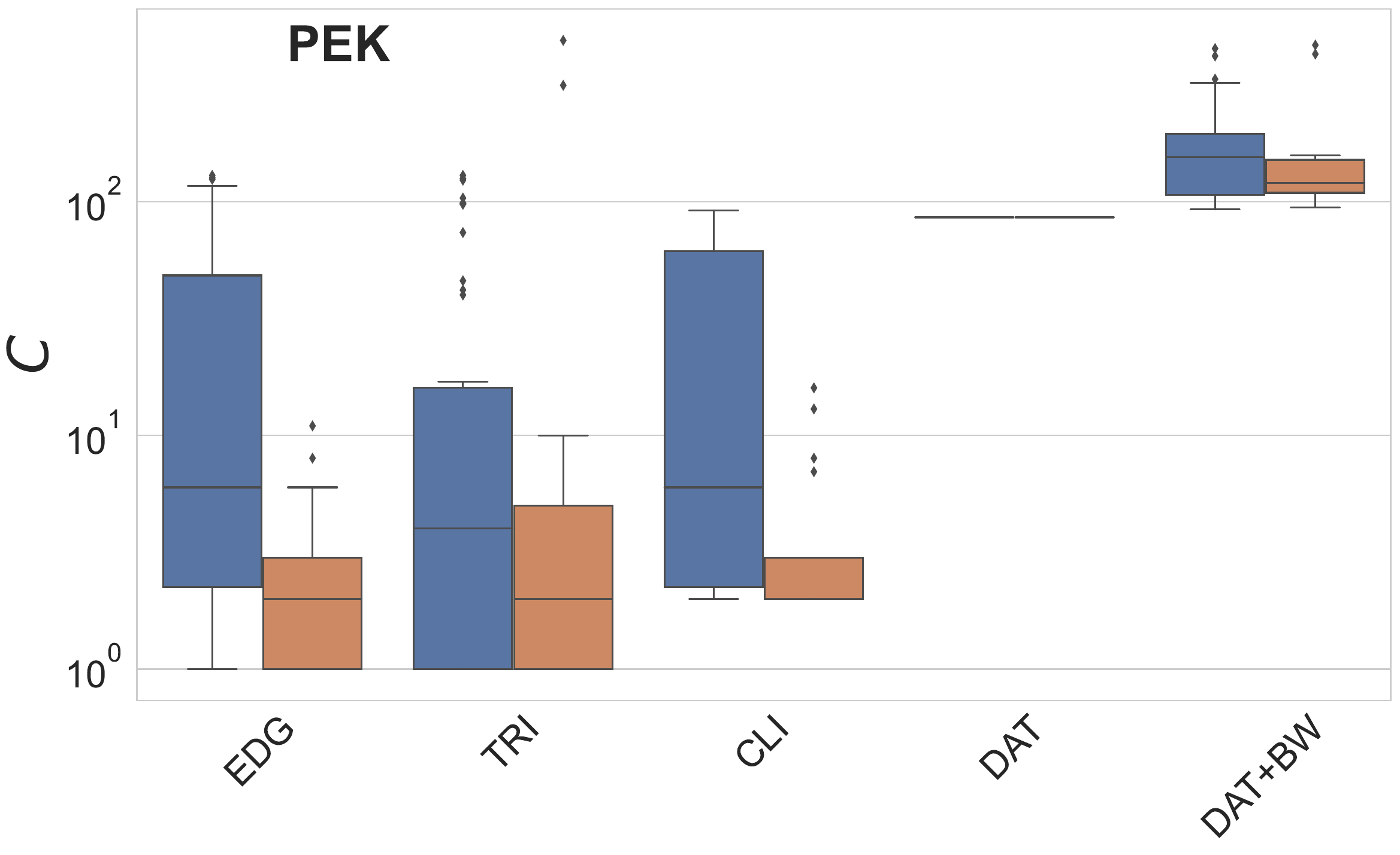}
   \end{minipage}
    \begin{minipage}[b]{0.49\textwidth}
        \includegraphics[width=.86\textwidth]{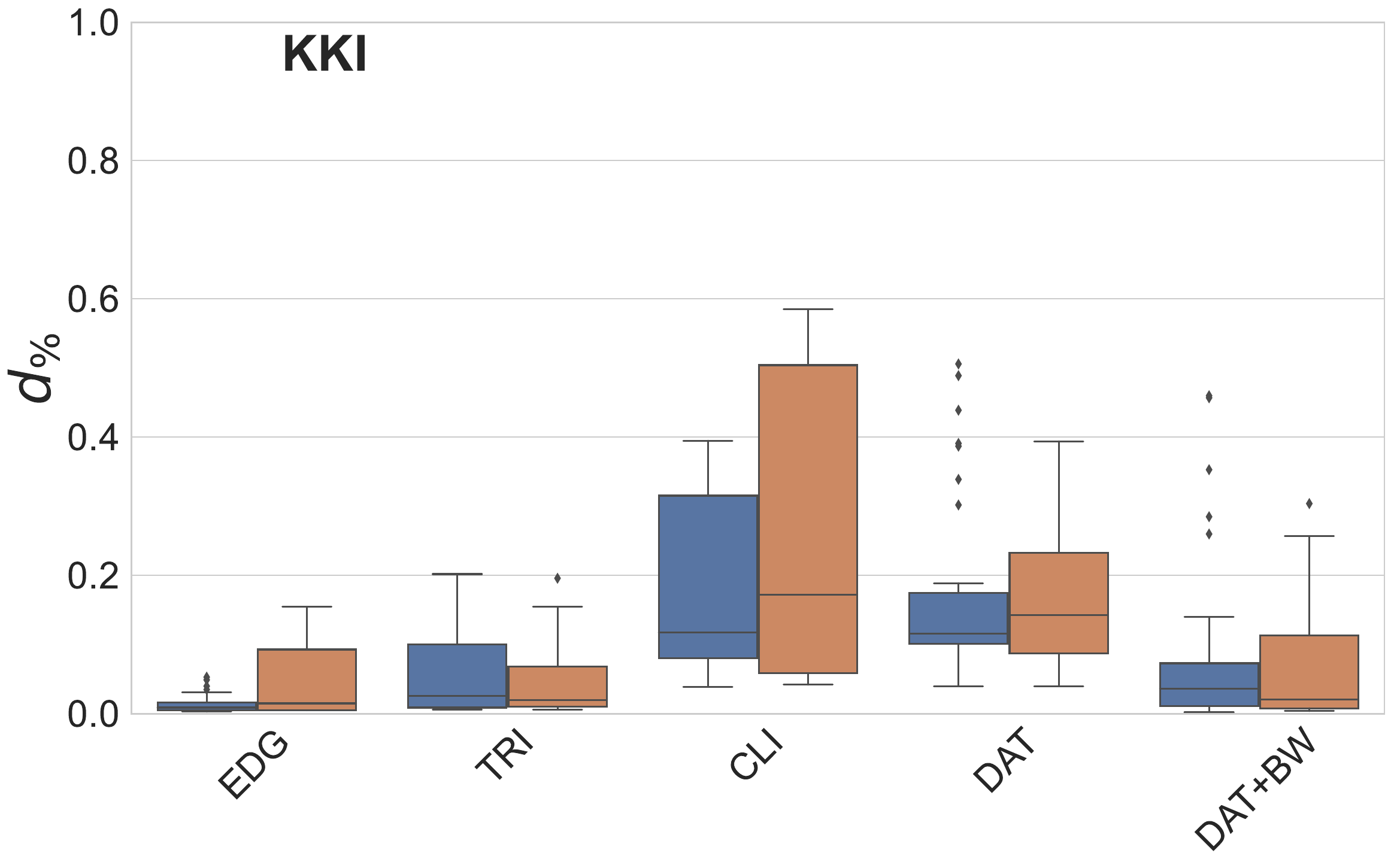}
   \end{minipage}
   \begin{minipage}[b]{0.49\textwidth}
        \includegraphics[width=.86\textwidth]{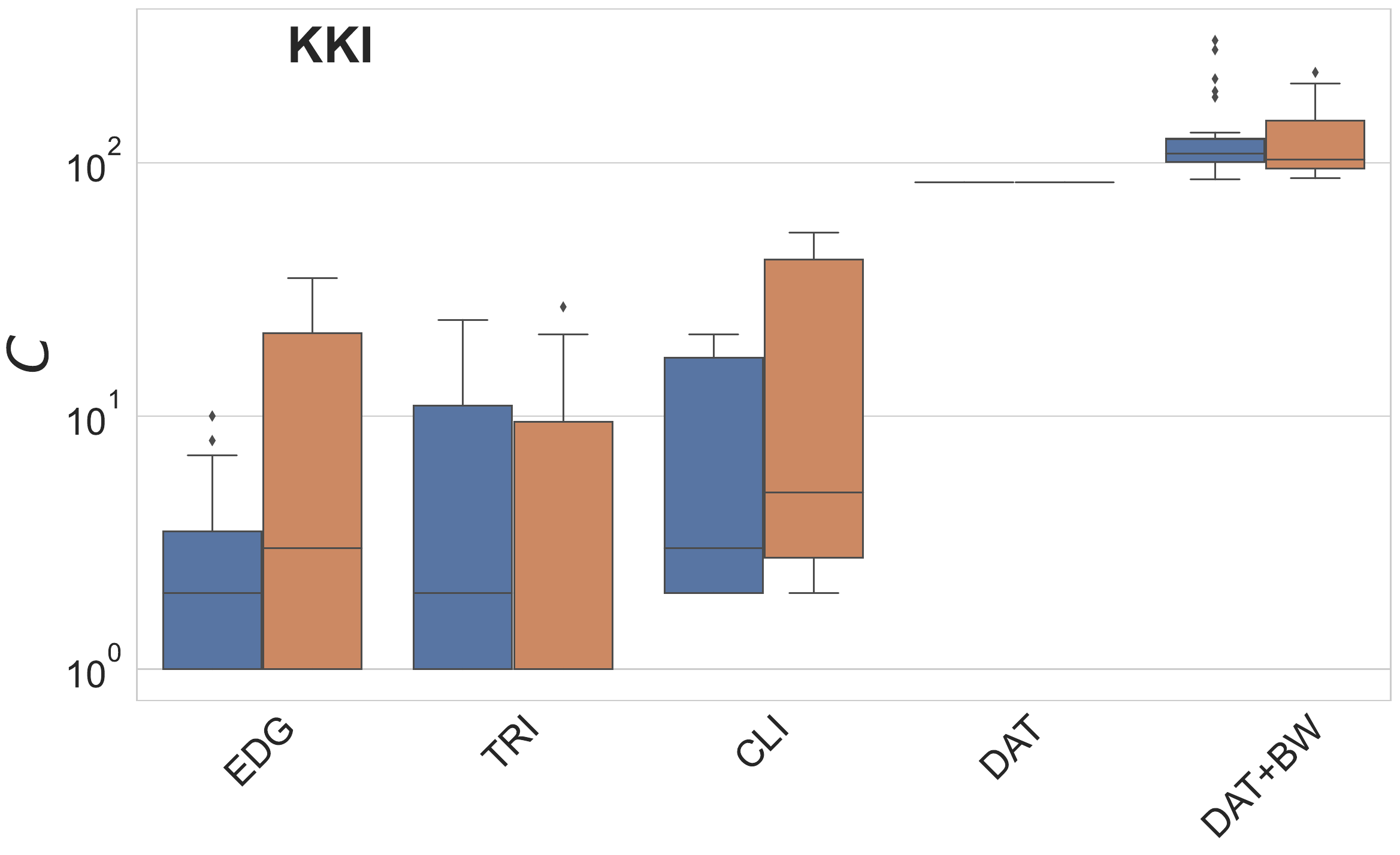}
   \end{minipage}
   \caption{Distribution of symmetric differences between original and counterfactual graphs ($d_{\%}$) and of computation times ($I$), for each counterfactual generation method, for ADHD (1st), OHSU (2nd), PEK (3rd), and KKI (4th).}
   \label{fig:distance_methods_app}
\end{figure}

\Cref{fig:distance_methods_app} shows the distribution of symmetric differences between original and counterfactual graphs ($d_{\%}$) and of computation times (\textit{I}), for each method, for ADHDF (1st row), OHSU (2nd row), PEK (3rd row), and KKI (4th row).

\clearpage

\begin{figure}[h]
\centering
    \begin{minipage}[b]{\textwidth}
    \centering
    \includegraphics[clip,width=.95\textwidth]{Figures/brains/legend.pdf}
    \end{minipage}
    \begin{minipage}[b]{0.67\textwidth}
    \centering
    \includegraphics[width=.86\textwidth]{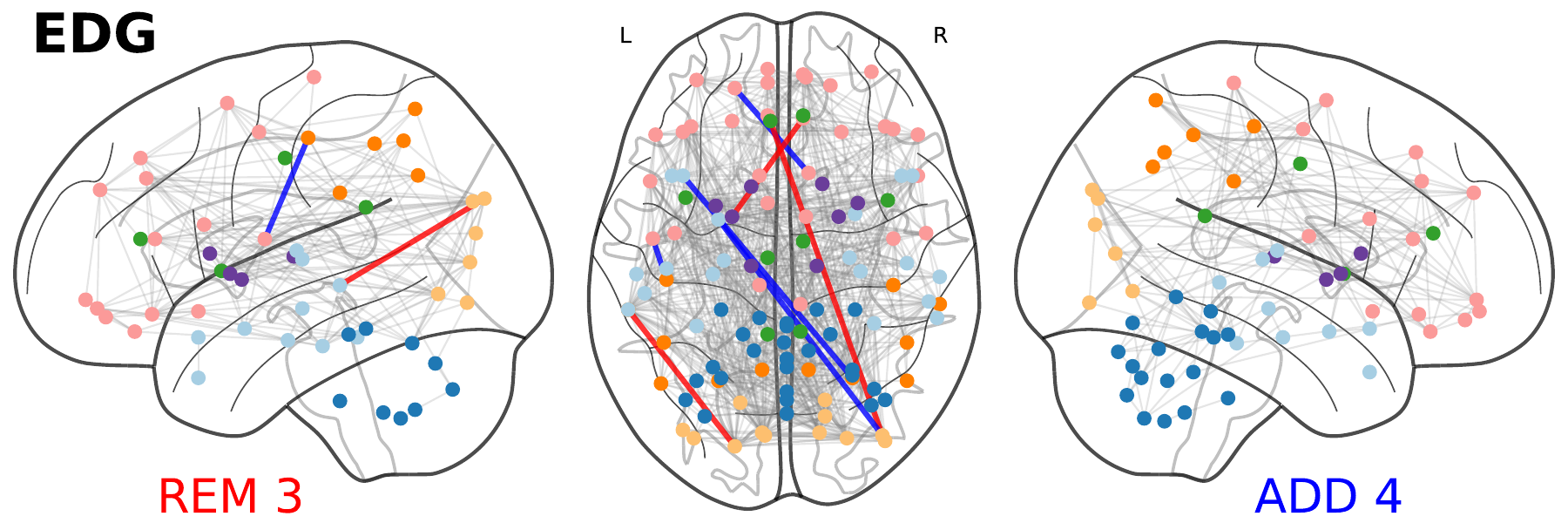}
   \end{minipage}
   \hfill
   \begin{minipage}[b]{0.32\textwidth}
    \centering
    \includegraphics[width=.8\textwidth]{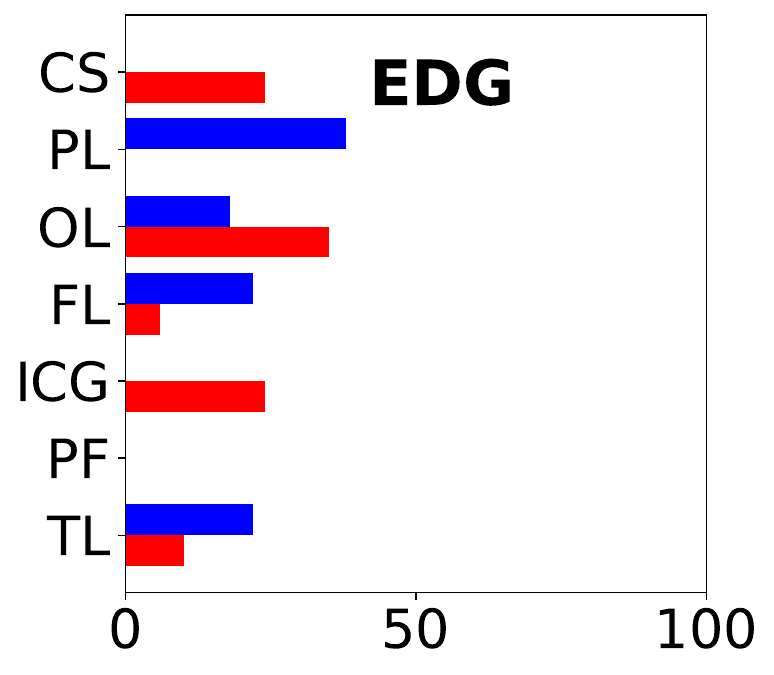}
   \end{minipage}
   \begin{minipage}[b]{0.67\textwidth}
    \centering
    \includegraphics[width=.86\textwidth]{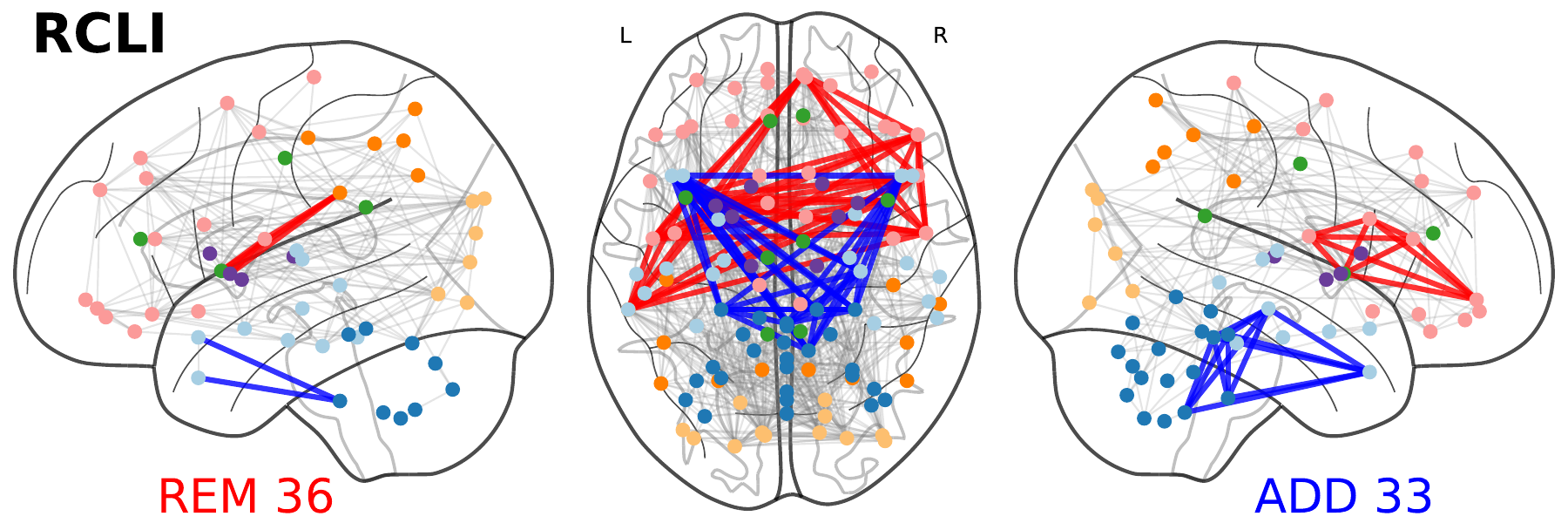}
   \end{minipage}
   \hfill
   \begin{minipage}[b]{0.32\textwidth}
    \centering
    \includegraphics[width=.8\textwidth]{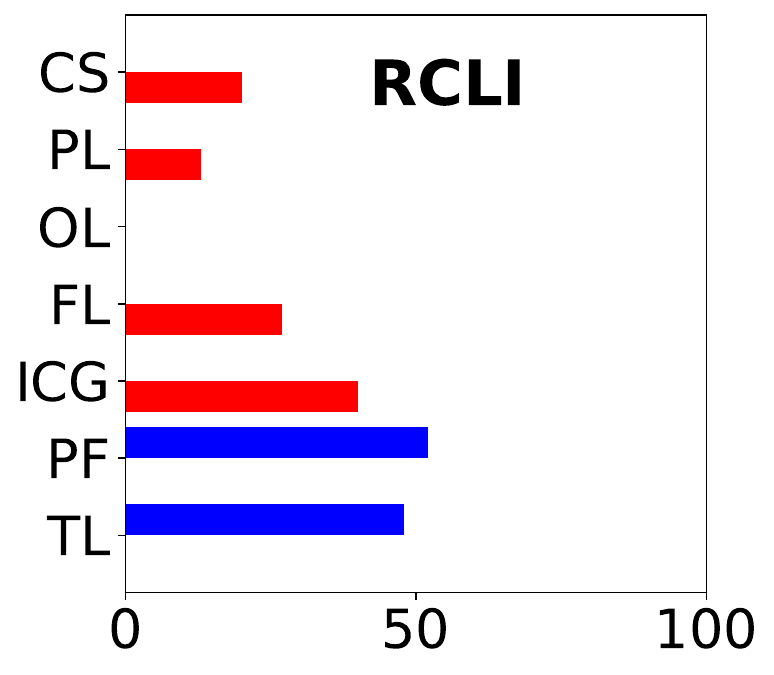}
   \end{minipage}
   \begin{minipage}[b]{0.67\textwidth}
    \centering
    \includegraphics[width=.86\textwidth]{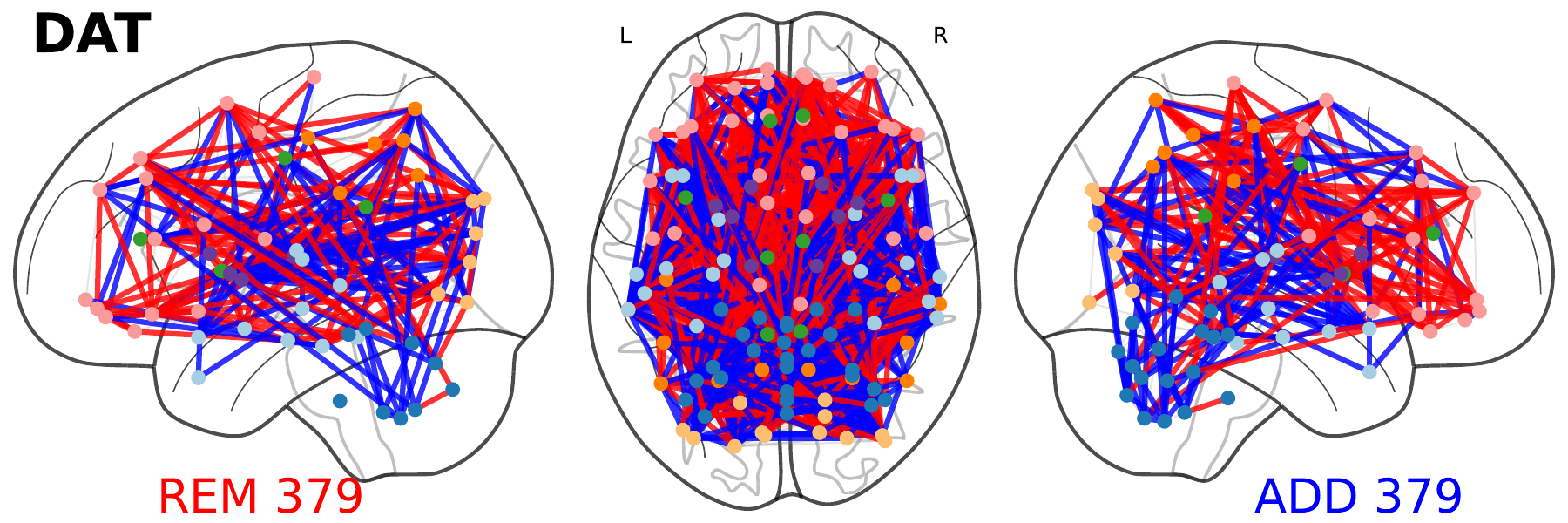}
   \end{minipage}
   \hfill
   \begin{minipage}[b]{0.32\textwidth}
    \centering
    \includegraphics[width=.8\textwidth]{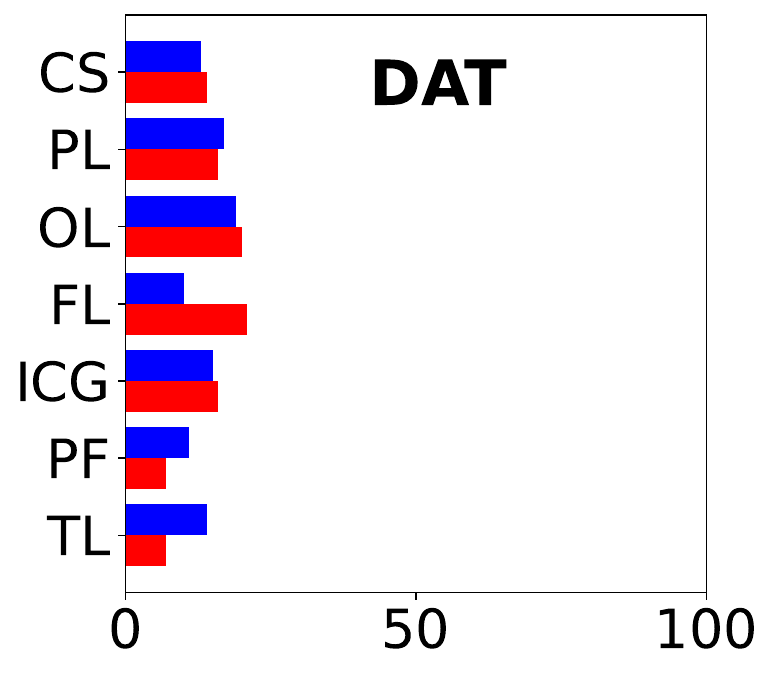}
   \end{minipage}
   \begin{minipage}[b]{0.67\textwidth}
    \centering
    \includegraphics[width=.86\textwidth]{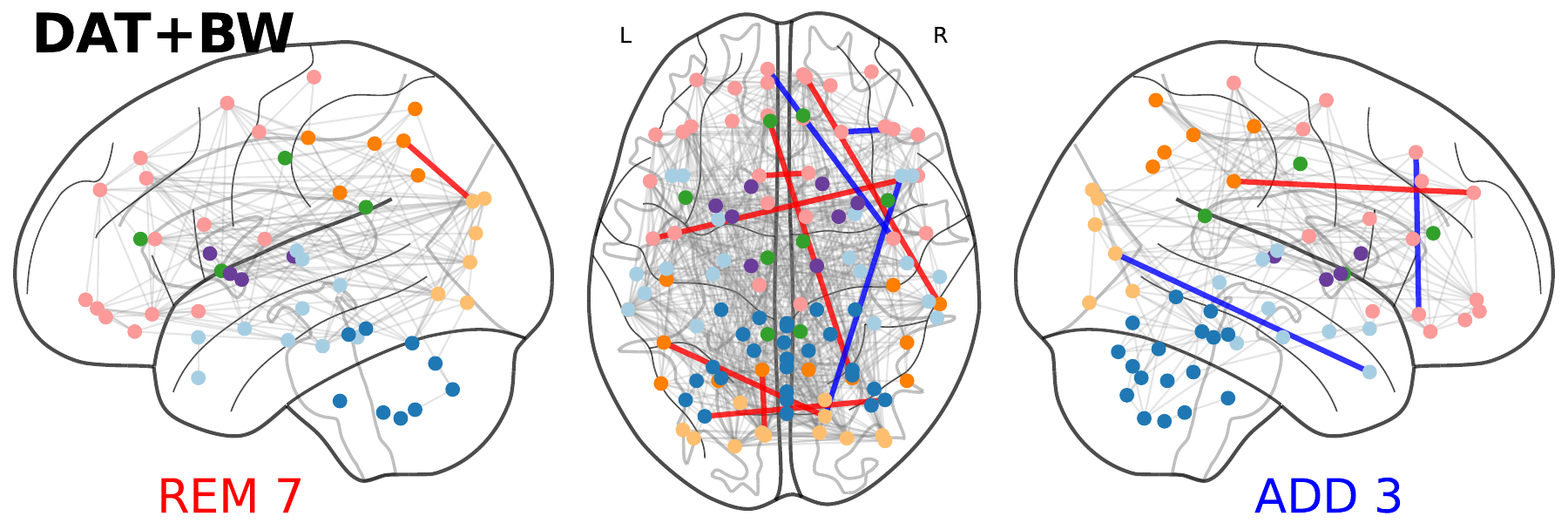}
   \end{minipage}
   \hfill
   \begin{minipage}[b]{0.32\textwidth}
    \centering
    \includegraphics[width=.8\textwidth]{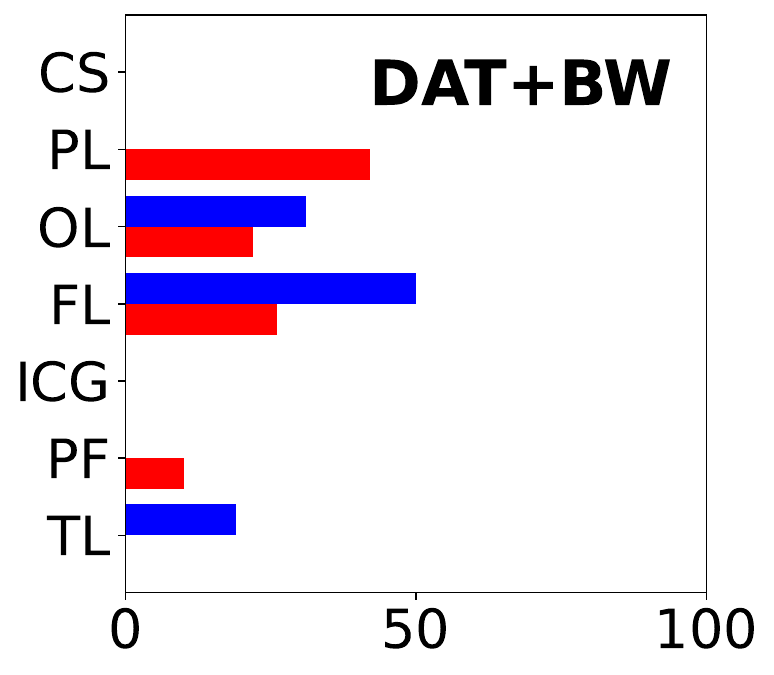}
   \end{minipage}
   \vspace{-6mm}
   \caption{Counterfactual graphs found by \edg, \rcli, \dataset, and \datasetbw, for patient $22$ in ADHD. On the left, the connectome of the patient with \textit{\color{red}connections to remove} (\emph{REM} is the number of edges removed), and \textit{\color{blue}connections to add} (\emph{ADD} is the number of edges added) to generate the counterfactual. On the right, to which brain lobes the ROIs of the edges added (in blue) and removed (in red) belong (as percentages).}
   \label{fig:example2_bis}
\end{figure}

\Cref{fig:example2_bis} compares the counterfactual graphs found by \edg, \rcli, \dataset, and \datasetbw, for patient $22$ in ADHD. The figure shows both the connectome of the patient with the changes highlighted in red (edges removed) and in blue (edges added), and the barplots indicating to which brain lobes the ROIs of the edges added (in blue) and removed (in red) belong (as percentages).

